\documentclass[journal]{IEEEtai}

\usepackage[colorlinks,urlcolor=blue,linkcolor=blue,citecolor=blue]{hyperref}

\usepackage{color,array}
\usepackage{cite}
\usepackage{amsmath,amssymb,amsfonts}
\usepackage{algorithmic}
\usepackage{textcomp}
\usepackage{url}
\usepackage{bm}
\usepackage{times}
\usepackage{epsfig}
\usepackage{graphicx}
\usepackage{amsmath}
\usepackage{multirow}
\usepackage{amssymb}
\usepackage{subfig}
\usepackage{graphicx}
\usepackage{float}
\usepackage{booktabs}
\usepackage{textcomp}
\usepackage{adjustbox}\usepackage{mathtools}
\usepackage{array}
\usepackage{makecell,booktabs}
\usepackage{multirow}
\usepackage{stackengine}
\usepackage{algorithm}
\usepackage{graphicx}
\usepackage[dvipsnames]{xcolor}
\usepackage{xcolor,colortbl}

\begin{document}

\title{CSNAS: Contrastive Self-supervised Learning Neural Architecture Search via Sequential Model-Based Optimization}
\author{{Nam Nguyen, \IEEEmembership{Student Member, IEEE},
J. Morris Chang, \IEEEmembership{Senior Member, IEEE}}
\thanks{Nam Nguyen is with the Electrical Engineering Department, University of South Florida, Tampa, Florida. (e-mail: namnguyen2@usf.edu)}}
\markboth{}
{Nam Nguyen \MakeLowercase{\textit{et al.}}: Contrastive Self-supervised Neural Architecture Search}


\maketitle

\begin{abstract}
This paper proposes a novel contrastive self-supervised neural architecture search algorithm (NAS), which completely alleviates the expensive costs of data labeling inherited from supervised learning. Our algorithm capitalizes on the effectiveness of self-supervised learning for image representations, which is an increasingly crucial topic of computer vision. First, using only a small amount of unlabeled train data under contrastive self-supervised learning (C-SSL) allow us to search on a more extensive search space, discovering better neural architectures without surging the computational resources. Second, we entirely relieve the cost for labeled data (by contrastive loss) in the search stage without compromising architectures' final performance in the evaluation phase. Finally, we tackle the inherent discrete search space of the NAS problem by sequential model-based optimization via the tree-parzen estimator (SMBO-TPE), enabling us to significantly reduce the computational expense response surface. An extensive number of experiments empirically show that our search algorithm can achieve state-of-the-art results with better efficiency in data labeling cost, searching time, and accuracy in final validation.
\end{abstract}

\begin{IEEEImpStatement}
Although transfer learning has achieved many computer vision tasks, finding a customized neural architecture for a specific dataset is still a promising solution for higher performance. However, the process of searching an optimal model usually demands extensive computational resources. This paper introduces a NAS algorithm that leverages newly developed C-SSL for image representations. Thus, our proposed approach entirely alleviates the cost of data labeling in the search stage. Moreover, our approach outperforms state-of-the-art NAS algorithms in mainstream datasets regarding predictive performance and computational expense.
\end{IEEEImpStatement}

\begin{IEEEkeywords}
Neural architecture search; Self-supervised learning; Sequential model-based optimization.
\end{IEEEkeywords}

\section{Introduction}
Automated neural search algorithms have significantly enhanced the performance of deep neural networks on computer vision tasks. These algorithms can be categorized into two subgroups: (1) flat search space, where automated methods attempt to fine-tune the choice of kernel size, the width (number of channels) or the depth (number of layers), and (2)  (hierarchical) \textit{cell-based search space}, where algorithmic solutions search for more minor components of architectures, called \textit{cells}. A single neural cell of deep models possesses a complex graph topology, which will be later stacked to form a more extensive network.

Although state-of-the-art NAS algorithms have achieved an increasing number of advances, several problematic factors should be considered. The main issue is that most NAS algorithms use the accuracy in validation inherited from supervised learning (SL) as the selection criteria. It leads to a computationally expensive search stage when it comes to sizable datasets. Hence, recent automated neural search algorithms usually did not search directly on large datasets but instead searching on a smaller dataset (CIFAR-10), then transferring the found architecture to more enormous datasets (ImageNet). Although the performance of transferability is remarkable, it is reasonable to believe that searching directly on source data may drive better neural solutions. Besides, entirely relying on SL requires the cost of data labels. It is not considered a problematic aspect in well-collected datasets, such as CIFAR-10 or ImageNet, where the labeled samples are adequate for studies. However, it may become a considerable obstacle for NAS when dealing with data scarcity scenarios. Take a medical image database as an example, where studies usually cope with expensive data curation, especially involving human experts for labeling. Hence, the remedy for such problems is vital to deal with domain-specific datasets, where the data curation is exceptionally costly. Finally, cell-based NAS algorithms' time complexity increases when the number of intermediate nodes within cells is ascended in the prior configurated search space. Previous works show that searching on larger space brings about better architectures. However, the trade-off between predictive performance and search resources is highly considerable.

We propose an automated cell-based NAS algorithm called Contrastive Self-supervised Neural Architecture Search (CSNAS). Our work's primary motivation is to offer high-performance models by expanding the search space without any trade-off of search cost. It is clear that searching larger space potentially increases the chance to discover better solutions. We realize that goal by employing the advances of self-supervised learning (SSL), which only requires a small number of samples used for the search stage to learn image representations. Besides, thanks to the nature of SSL, we entirely relieved the cost for labeled data in the search stage. It is significant to mention since, in many domain-specific computer vision tasks, the unlabeled data is abundant and inexpensive, while labeled samples are typically scarce and costly. Thus, we are now able to use a cost-efficient strategy for NAS. Furthermore, we directly address the natural discrete search space of the NAS problem by SMBO-TPE, which evaluates the costly contrastive loss by computationally inexpensive surrogates. We have been made our implementation for public availability, hoping that there will be more research investigating our algorithm's efficiency concerning computer vision applications. The code for implementation can be found at GitHub: "https://github.com/namnguyen0510/CSNAS". Moreover, the detailed hyper-parameter setting is given in Section~\ref{appendix:arcsearch} and~\ref{appendix:arceval}.

\section{Related Work}
In this section, we first introduced state-of-the-art supervised NAS in Section~\ref{relatedwork_snas}. Second, the overview of SSL will be discussed in Section~\ref{related_ssl}. We then outline state-of-the-art self-supervised NAS algorithm in Section~\ref{related_sslnas}. Finally, we provide our point-of-view on the merits of NAS algorithms in Section~\ref{merits}, followed by a highlight to distinguish our CSNAS from other approaches. The summary of contribution will also be given at the end of the section.
\subsection{Supervised Neural Architecture Search} \label{relatedwork_snas}

We briefly outline the advantages of state-of-the-art supervised NAS algorithms in this section. The discovered architectures searched by supervised NAS algorithms have established highly competitive benchmarks in both image classification tasks \cite{zoph2018learning, liu2017hierarchical, liu2018progressive, real2019regularized} and object detection \cite{zoph2018learning}. The best state-of-the-art supervised NAS algorithms are extremely computationally expensive despite their remarkable results. A reason for inefficient searching process is due to the dominant approaches: reinforcement learning \cite{zoph2016neural}, evolutionary algorithms \cite{real2019regularized}, sequential model-based optimization (SMBO) \cite{liu2018progressive}, MCTS \cite{negrinho2017deeparchitect} and Bayesian optimization \cite{kandasamy2018neural}. For instance, searching for state-of-the-art models took $2000$ GPU days under reinforcement learning framework \cite{zoph2016neural}, while evolutionary NAS required $3150$ GPU days \cite{real2019regularized}. The main reason for such extreme computational expense can be attributed to the selection of dataset for the search phase. In the pioneers NAS algorithms, the dataset used for both search and evaluation phase is typically ImageNet\cite{deng2009imagenet}, which includes a considerable large number of samples. As a consequence, the computational burden for the search phase is tremendously massive. Directly tackling this issue, following NAS algorithms introduce the usage of proxy dataset for the search phase in order to reduce the computational expense. The common choice for proxy of ImageNet is CIFAR-10\cite{krizhevsky2009learning} in the NAS-related study. In particular, we search the optimal models on CIFAR-10 and investigate the transferability to ImageNet. As the result, the search cost for NAS algorithms significantly reduced to an affordable GPU hours. For example, RelativeNAS\cite{tan2021relativenas} introduce an efficient population-based NAS algorithm which can discover high predictive performance model within $0.4$ GPU days. Moreover, several well-established algorithmic solutions for NAS using proxy dataset for the search phase have overcome expensive computational requirements without the lack of scalability: differentiable architecture search \cite{liu2018darts,xu2019pc,chen2021progressive} enables gradient-based search by using continuous relaxation and bilevel optimization; progressive neural architecture search utilized heuristic search to discover the structure of cells \cite{liu2018progressive}; sharing or inheriting weights across multiple child architectures \cite{elsken2017simple, bender2018understanding, pham2018efficient, cai2018efficient} and predicting performance or weighting individual architecture \cite{baker2017accelerating, brock2017smash}. Although these latter approaches can reach state-of-the-art results with efficiency concerning searching time, they may be affected by the inherited issue from gradient-based approach, which finds the local minimum solution. Apart from that, most state-of-the-art NAS algorithms require full knowledge of training data. Thus, the searching process is frequently performed on a smaller dataset (e.g., CIFAR-10 or CIFAR-100), then the discovered architecture will be trained on a bigger dataset (e.g., ImageNet) to evaluate the transferability. Although we can relax the constraint of using the same dataset in the search phase and evaluation phase by transferring the learned architecture, it is reasonable to believe that training with the constrain may result in better solutions.

\subsection{Self-supervised Learning for Learning Image Representation}\label{related_ssl}
\begin{figure}[t]
    \centering
    \includegraphics[width = 0.46\textwidth]{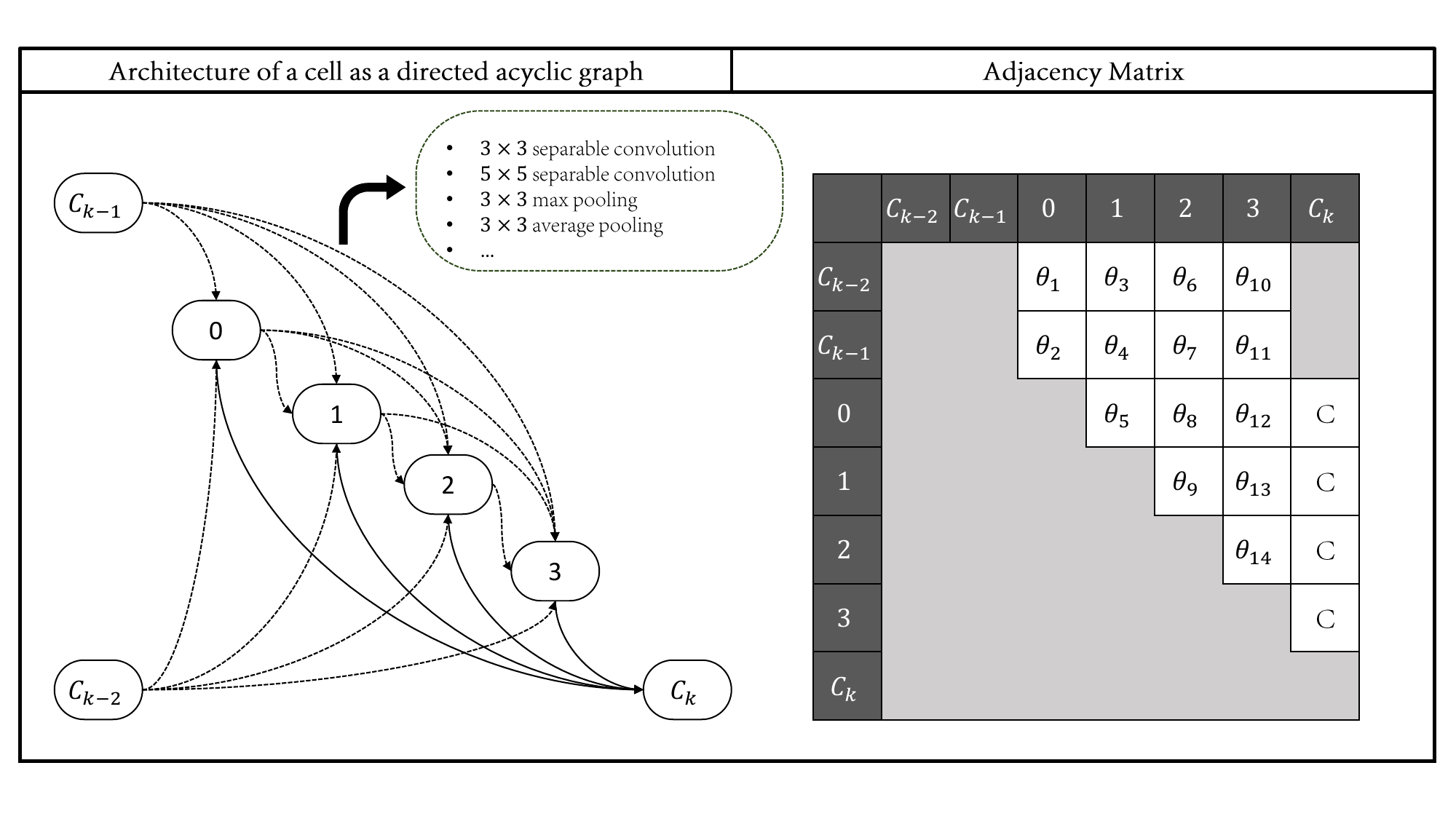}
    \caption{A graph representation of cell architecture. Left figure: dashed lines represent connections between nodes via a choice of operations; solid lines depict fixed connections. Right figure: the adjacency matrix of the corresponding cell architecture. All shaded entries are zeros since it is impossible to establish corresponding connections. Each $\theta_i$ is a random variable representing a choice of operation.}
    \label{cellconstruction}
\end{figure}
SSL has established extremely remarkable achievements in natural language processing \cite{mikolov2013word2vec, joulin2016fasttext, devlin2018bert}. Hence learning visual representation has drawn great attention with a large scale of literature that has explored the application of SSL for video-based and image-based classification. Within the scope of this paper, we only focus on SSL for an image classification task. Generally, mainstream approaches for learning image representations can be categorized into two classes: generative, where input pixels are generated or modeled \cite{pathak2016context, zhang2017split, radford2015unsupervised, donahue2016adversarial}; and discriminative, where networks are trained on a pretext task with a similar loss function. The key idea of discriminative learning for visual representations is the designation of pretext task, which is created by pre-assigned targets and inputs derived from unlabeled data: distortion \cite{dosovitskiy2015discriminative}, rotation \cite{gidaris2018unsupervised}, patches or jigsaws \cite{doersch2015unsupervised}, colorization \cite{zhang2016colorful}. Moreover, the most recent research interests of SSL have been drawn from contrastive learning in the input latent space \cite{oord2018representation, henaff2019data, srinivas2020curl, grill2020bootstrap, chen2020simple}, which promisingly showed comparable achievement to SL. \cite{chen2020simple, grill2020bootstrap,misra2020self}
provide convincing evidence of the capacity to learn image representations of C-SSL, which only require a small proportion of training data (1\% to 10\%) on linear evaluation to reach SL performance. This work also studied the effect of C-SSL of different backbone architectures, which empirically shows that the predictive performance consistently gains when scaling backbone networks. Moreover, a solid theoretical analysis of SSL (or self-training) is introduced by \cite{wei2020theoretical}. Under simplified but practical assumptions that (1) a subset of low confident samples must expand to a neighborhood with higher confidence concerning the subset and (2) neighborhoods of samples from different classes are minimally overlapped, SSL on unlabeled data with input-consistency regularization will result in higher accuracy in comparison to labeled data.

\subsection{Self-supervised Learning Neural Architecture Search}
\label{related_sslnas}
First, we would like to give a simplified design for self-supervised NAS algorithms, including two main components: (1) choices of the SSL method and (2) search strategy. In the current literature, SSNAS \cite{kaplan2020self} and UnNAS \cite{liu2020labels} also introduced self-supervised NAS algorithms that leverage the DART algorithm for search strategy. Moreover, the choice of the SSL method of SSNAS is SimCRL, while UnNAS utilized invariant pretext tasks. It is worth mentioning that SSL based on pretext tasks is distinguishable from C-SSL SimCRL. The former approach assigns pretext task labels to unlabeled data (rotation angle, colorization type, suffer and solve jigsaw puzzles), then trains the model based on these generated labels under supervised learning using the conventional loss function of supervised learning. To some extent, we still observe supervised learning within SSL based on pretext tasks since it can be considered as supervised training. On the other hand, C-SSL trains to maximize the agreement between positive pairs (augmented views from the same instance) while minimizing the agreement between negative pairs (augmented view from different instances). Hence, C-SSL eliminates the trait of supervised learning by noise contrastive loss while achieving much better results (very closed to supervised learning) in comparison to pretext task SSL\cite{chen2020simple,misra2020self,grill2020bootstrap}. 

SSNAS is inspired by the fact that all architectures in the search space are over-fitted on the training data. Thus, their selection criteria are based on the architecture's generalization over the training data. To realize their goal, they used the margin-based search, which involves two splits of training data. Each candidate is trained on one split to increase the margin between samples. The selected candidate is the model that maintains the most significant distance between instances. 

On the other hand, UnNAS attempted to answer a fascinating question: "Are labels necessary for NAS?". The work searched neural architecture by DART algorithm using the pretext-task-assigned dataset, including rotation, colorization, and solving a jigsaw puzzle. Both studies provide compelling experimental results showing that self-supervised NAS can search for high-performance neural solutions while relieving the cost of data annotations in the search stage.

\subsection{Merits of NAS algorithms and Contribution of CSNAS}
\label{merits}
In this section, we will discuss the merits and evaluation criteria for NAS algorithms. 
Early supervised NAS algorithms \cite{zoph2016neural,real2019regularized} are several of frontiers in supervised NAS algorithms, which successfully searched models while remaining the constraint of searching data and evaluating data. However, they need to make a massive trade-off with thousands of searching hours. Thus, such an approach's disadvantage is that it is nearly impossible to implement with limited computational resources. With the growth of research interest from the field, many sub-sequence works successfully reduce the computational requirements while maintaining high predictive power \cite{liu2018darts,chen2019progressive,xie2018snas,pham2018efficient}. However, such algorithms still have several weaknesses. For example, DART and sub-sequence approaches such as PC-DART and P-DART result in different cell architectures even though they share the same initial search space. Thus, DART could find a local optimum for the NAS problem. However, we cannot deny that DARTS offers us the first efficient gradient-based NAS algorithms in terms of accuracy and searching resources. 

Moreover, recent work such as SSNAS and UnNAS successfully relieve the cost of labeled data in the search stage while achieves comparable results with supervised NAS. Any supervised NAS algorithm can perform self-supervised NAS if we take the labels for grated in the search stage. Hence, it is not reasonable to compare supervised NAS and self-supervised NAS based on the test accuracy. Intuitively, supervised NAS algorithms should achieve a better predictive performance since they have access to the train set's full knowledge, which includes data with annotations. However, we are more often than not dealing with computer vision problems involving data scarcity scenarios, in which the cost for data annotations is expensive while unlabeled data are much more abundant. In this case, self-supervised NAS algorithms surpass every supervised NAS algorithm since they can leverage the additional unlabeled samples, which are entirely take for granted by supervised NAS algorithms.  Nevertheless, self-supervised NAS also inherits several issues from its backbone training procedure. First, there could be a loss of information while using pretext tasks labels or contrastive noise estimators compared to the conventional loss of supervised learning. Recent self-supervised NAS algorithms UnNAS and SSNAS show a minimal gap of such loss on a nicely collected database (CIFAR-10 and ImageNet), including well-represented training samples with equal distribution. However, such loss of information is challenging to indicate since it is data-dependent universally. We can approximate the loss of information by comparing supervised NAS algorithms on a specific database. Second, the time complexity of search under C-SSL is more considerable than supervised learning search since it requires multiple inputs to evaluate the contrastive loss. This issue does not appear in pretext task SSL from UnNAS since it trains a given neural architecture under supervised learning with labels generated by given pretext tasks. However, the increase in the time complexity due to C-SSL is extremely minor in comparison to the whole time complexity of NAS algorithms, since we only need to perform back-propagation on a sole neural candidate

We distinguish our proposed CSNAS from other self-supervised NAS algorithms by two main points. First, we leverage a different C-SSL PIRL\cite{misra2020self} for learning image representations within the search phase. SSNAS used SimCRL, which benefits from an extensive training batch size (initially 4096 samples per mini-batch) and a long searching process. Thus, it is a clear obstacle for practical purposes due to its computationally expensive resources. In contrast, by using a memory-bank, PIRL allows SSL with computationally affordable batch size while remaining the adequate negative samples per mini-batch ($4\times$ smaller compared to SimCRL). Second, it is clear that there will be an increase in the searching time if we directly employ supervised NAS's search strategy for self-supervised NAS. Hence, we compromise the surge by sequential model-based optimization (SMBO) with a tree-structured Parzen estimator (TPE). Our empirical experiments (Sect~\ref{arceval}) show that architectures searched by CSNAS on CIFAR-10 outperform hand-crafted architectures \cite{huang2017densely, nokland2019training,he2016identity} and can achieve high predictive performance in comparison to state-of-the-art NAS algorithms. We also investigate our proposed approach's domain adaption on a medical imaging database under a data scarcity scenario. The case study provides us a deeper insight into the effectiveness of contrastive learning and gives us a promising result of CSNAS in practice. We summarize our contributions as follows:

\begin{itemize}
    \item We introduce a novel algorithm for self-supervised NAS. C-SSL enables efficient search with access to a small proportion of data without annotations. Thus, our proposed approach relieves the computational cost for the search stage, including searching time and labeling cost.
    \item Our approach is the first NAS framework based on Tree Parzen Estimator (TPE) \cite{bergstra2011algorithms}, which is well-designed for discrete search spaces of cell-based NAS. Moreover, the prior distribution in TPE is non-parametric densities, which allows us to sample many candidates to evaluate the expected improvement for surrogates, which is computationally efficient. Moreover, surrogate models' usage reduces the expensive cost of true loss function during the search phase. Thus, we can improve search efficiency and accuracy with CSNAS even when the search space is expanded.  CSNAS achieves state-of-the-art results with $2.66\%$ and $25.6\%$ test error in CIFAR-10 \cite{krizhevsky2009learning}  and ImageNet  \cite{imagenet_cvpr09}, without any trade-off of search costs. 
    \item We also evaluate the domain adaption ability of our proposed approach on a medical imaging database involving skin lesion classification. Our searched models on a limited dataset without annotations is well-designed and well-customized for skin lesion classification, which outperforms state-of-the-art architectures under transfer learning, reaching $88.68\%$ of accuracy in the test set. 
\end{itemize}

We organize our work as follows: Sect.~\ref{methods} mathematically and algorithmically illustrates our proposed approach, while Sect.~\ref{experiment} will give the experimental results and comparison with state-of-the-art NAS on CIFAR-10 and ImageNet. Moreover, we report a detailed analysis of a case study involving skin lesions in Section~\ref{case_study}. Finally, we will discuss the analysis and future work in Sect.~\ref{conclusion}. 

\section{Methodology}\label{methods}
We will generally describe two main fundamental components of our study: neural architecture search (NAS) and contrastive self-supervised visual representation learning (C-SSL) in Sect.~\ref{NAS} and  Sect.~\ref{SSL}. Finally, we will formulate the Contrastive Self-supervised Neural Architecture Search (CSNAS) as a hyper-parameters optimization problem and establish its solution by tree-structured Parzen estimator in Sect.~\ref{TPE_section}

\subsection{Neural Architecture Search}  \label{NAS}
\subsubsection{Neural Architecture Construction \label{NAconstruction}}

We employ the architecture construction from \cite{zoph2016neural,zoph2018learning}, where searched cell are stacked to form the final convolutional network. Each cell can be represented as a directed acyclic graph of $N + 3$ nodes, which are the feature maps and each corresponding operation $o^{(i,j)}$ forms directed edges $(i,j)$. Following \cite{zoph2018learning, liu2017hierarchical, liu2018progressive, real2019regularized}, we assume that a single cell consists of two inputs (outputs of the two previous layers $C_{k-2}$ and $C_{k-1}$), one single output node $C_{k}$ and $N$ intermediate nodes. Latent representations in intermediate nodes are included, which computed as in \cite{liu2018darts}:
$$\bm{x}^{(j)} = \sum_{i<j}o^{(i,j)}(\bm{x}^{i})$$
The generating set $\mathcal{O} = \{ o^{(i,j)}\}$ for operations between nodes is employed from the most selected operators in \cite{zoph2016neural,zoph2018learning,real2019regularized}, which includes seven non-zero operations: $3 \times 3$ and $5\times 5$ dilated separable convolution, $3 \times 3$ and $5\times 5$ separable convolution, $3 \times 3$ max pooling and average pooling, identity and zero operation (skip-connection). We retain the same search space as in DARTS \cite{liu2018darts}.
The total number of possible DAGs (without graph isomorphism) containing $N$ intermediate nodes with a set of operators $\mathcal{O}$ are:
$$\prod_{k=1}^{N} \frac{(k+1)k}{2} \times (|\mathcal{O}|^2).$$ We encode each cell's structure as a configurable vector $\bm{\theta}$ of length $\sum_{i=2}^{N+1} i$ and simultaneously search for both normal and reduction cells. Therefore the total number of viable cells will be raised to the power of 2. Thus, the cardinality of $\mathcal{F}$ - set of all possible configurations - is  $\big( |\mathcal{O}|^{\sum_{i=2}^{N+1} i} \big)^2.$ Also, we observed that the discrete search space for each type of cells (normal and reduction) is enormously expanded by a factor of $|\mathcal{O}|^{(N+1)}$ when increasing the number of intermediate nodes from $N$ to $N+1$. Consequently, state-of-the-art NAS can only achieve a low search time (in days) when using $N = 4$, while ascending $N$ to 5 usually takes a much longer search time, up to hundreds of days. Our approach treats the high time complexity of search space expansion by (1) using only a small proportion of data under contrastive learning for visual representations and (2) evaluating the loss by surrogate models, which requires much less computational expense. Details of these methods will be discussed in the next sections.
\subsubsection{Evaluation Criteria for NAS algorithms}

The evaluation metric for NAS algorithms considering three components, which are (1) versatility in different data scenarios, (2) computational expense for the search phase and (3) the power of representation learning from discovered models. \textbf{First}, self-supervised NAS algorithms completely relieves data annotations for the search phase, enabling a reduced cost in terms of data curation. \textbf{Second}, a good NAS algorithm should require affordable computational resources, which results in a reasonable computational cost for the search phase. \textbf{Finally}, a good NAS algorithm should derive a model with high representation learning ability with reasonable model complexity in term of number of parameters. The representations produced by searched architecture should have a high level of generalization. In other words, a intermediate-sized representations can capture complex concepts from an extensive number of inputs, while disentangle the variation factor and mitigate the variance in the data distribution. Although evaluation of representation learning ability remains as open question and mainly relies on the training task\cite{bengio2013representation}, we assume a simple but practical assumption that is a good NAS algorithm should result in a high performance model, in term of accuracy in validation. In particular, the validation accuracy is given as:
\begin{equation}
    \text{Test Error} = \bigg(1 - \frac{(\text{TP} + \text{TN})}{(\text{TP} + \text{TN} + \text{FP} + \text{FN})} \bigg)*100\%.
\end{equation}

\subsection{Contrastive Self-supervised Learning} \label{SSL}
\begin{figure}[t]
    \centering
    \includegraphics[width = 0.48\textwidth]{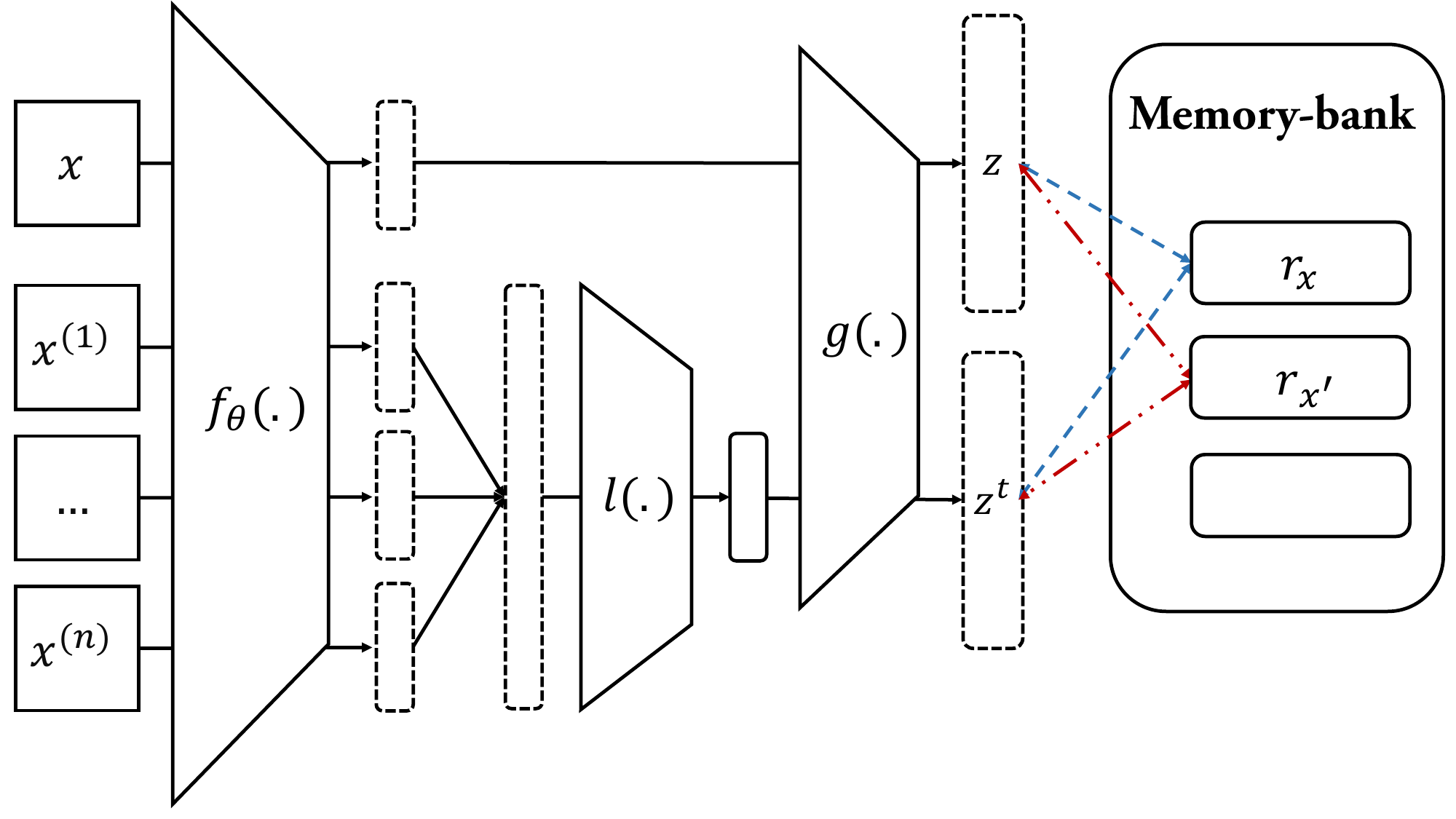}
    \caption{\textbf{PIRL genetic framework:} Input image along with its augmented views are fed forward the same neural candidate parameterized by a collection of model's weight $\bm{\theta}$. The derived feature maps are then projected onto lower dimensional representations by $l(.)$ and $g(.)$. Contrastive loss maximizes the similarity between the image representation of original image $x$ and its augmented views with its visual presentations $r_x$ in the memory-bank (dashed green line), while the agreement with negative samples $r_{x}^{'}$ is minimized (dashed-dotted red line). The usage of memory-bank enables us to cache all representations of all samples in the input data, which is the exponential moving average of representations from prior epochs.}
    \label{fig:SSL}
\end{figure}
We employ a recent contrastive learning framework in PIRL \cite{misra2020self}, allowing multiple views on a sample. 
Let $\mathcal{D} = \{\bm{x}_i \}_{i=1}^{i = |\mathcal{D}|}$ be a train set (for searching) and $\mathcal{F} = \{ f_{\mathbf{\bm{\theta}}} (.)\}_{\bm{\theta} \in \Theta}$ be a set of all candidates:
\begin{itemize}
    \item Each sample $\bm{x}_i \in \mathbb{R}^{H \times W \times 3}$ is first taken as input of a stochastic data augmentation module, which results in a set of correlated views $\bm{x}^t_i = \{\bm{x}_i^{(1)}, ... , \bm{x}_i^{(M)} \}$. Within the scope of this study, three simple image augmentations are applied sequentially for each data sample, including: random/center cropping, random vertical/horizontal flipping and random color distortions to grayscale. The set $\mathcal{X} = \{\bm{x}_{i}\} \cup \bm{x}^t_{i}$ is called positive pair of sample $\bm{x}_i$.
    
    \item Each candidate architecture $f_{\bm{\theta}}(.)$ is used as a base encoder to extract visual representations from both original sample $\bm{x}_i$ and its augmented views $\bm{x}^t_i$. We use the same multilayer perceptron $g(.)$ at the last of all neural candidates, projecting its feature maps of original $\bm{x}_i$ image under $f_{\bm{\theta}}(.)$ into a vector  $\mathbb{R}^p$. For augmented views, another intermediate MLP $l(.): \mathbb{R}^{Mp} \rightarrow \mathbb{R}^p$ is applied on concatenation of $\{f_{\bm{\theta}}(\bm{x}_{i}^{(m)})\}$ for ${m \in M}$. We denote the representing vectors of original image and augmented views as $\bm{z}_i$ and $\bm{z}^t_i$, respectively.
    
    
\end{itemize}

We also use the cosine similarity as the similarity measurement as in \cite{chen2020simple, misra2020self}, yielding $s(\bm{u}, \bm{v}) =\bm{u}^{T}\bm{v}/||\bm{u}||.||\bm{v}||$. Each minibatch of $K$ instances is randomly sampled from $\mathcal{D}$, giving $M\times K$ data points. Similar to \cite{chen2017sampling}, a positive pair is corresponding to in-batch negative examples, which are other $M(K-1)$ augmented samples, forming a set of negative sample $\mathcal{X'}$. Similarly, each negative sample is extracted visual representations as $\bm{z'}_i = g(f_{\bm{\theta}}(\bm{x'}_i))$. Following \cite{misra2020self}, we compute the noise contrastive estimator (NCE) of a positive pair $\bm{x}_i$ and $\bm{x}^t_i$ using their corresponding $\bm{z}_i$ and $\bm{z}^t_i$, given by
\begin{equation}\label{nce}
    \ell (\bm{z}_i,\bm{z}^t_i) =  \frac{\exp (\frac{s(\bm{z}_i, \bm{z}^t_i)}{\tau} )}{\exp (\frac{s(\bm{z}_i, \bm{z}^t_i)}{\tau} ) + \sum_{\bm{x'}_i \in \mathcal{X'}} \exp (\frac{s(\bm{z}_i^t, \bm{z'}_i)}{\tau})}.
\end{equation}
The estimators are used to minimize the loss:
\begin{equation}\label{closs}
    \mathcal{L}_{\text{NCE}}(\bm{x}_i, \bm{x}^t_i) = -\log \ell(\bm{z}_i, \bm{z}^t_i) - \sum_{\bm{x'} \in \mathcal{X'}} \log \big[ 1 - \ell \big( \bm{z}_i^t, \bm{z'}_i \big) \big]
\end{equation}

The NCE loss maximizes the agreement between the visual representation of the original image $\bm{x}_i$ and its augmented views $\bm{x}^t_i$, together with minimizing the agreement between $\bm{x}_i$ and $\bm{x'}_i$. We use memory-bank approach in \cite{misra2020self,he2020momentum} to cache the representations of all samples in $\mathcal{D}$. The representation $\bm{r}_{\bm{x}}$ in memory-bank $\mathcal{M}$ is the exponential moving average of $\bm{z_i}$ from prior epochs. The final objective function for each neural candidate is a convex function of two losses as in Equation~\ref{closs}:
\begin{equation}\label{final_loss}
    \mathcal{L}_{\bm{\theta}}(\bm{x}, \bm{x}^t) = \lambda \mathcal{L}_{\text{NCE}}(\bm{r}_{\bm{x}}, \bm{z}^t) + (1-\lambda)\mathcal{L}_{\text{NCE}}(\bm{r}_{\bm{x}}, \bm{z})
\end{equation}

Finally, these loss values establish the scoring criteria for models under sequential model-based optimization, which will be discussed in the next section.

\subsection{Tree-structured Parzen Estimator} \label{TPE_section}

As mentioned in the previous sections, we aim to search on a larger space in order to discover a better neural solution. Nevertheless, the main difficulty when expanding the search space is due to the exponential surge in the time complexity. Thus, we are motivated to study an optimization strategy that might reduce the computational cost. We employ the sequential model-based optimization (SMBO), which has been widely used when the fitness evaluation is expensive. This optimization algorithm can be a promising approach for cell-based NAS, since current state-of-the-art NAS algorithms use the loss in validation as the fitness function, which is computational expensive. The evaluation time for each neural candidate tremendously surges when the number of training samples or sample's resolution increases. In the current literature, PNAS \cite{liu2018progressive} is the first framework which apply SMBO for cell-based NAS. The surrogate model of in PNAS is used to predict model performance without training them. In contrast to their approach, where the fitness function is in-validation accuracy, we model the contrastive loss in Equation~\ref{final_loss} by a surrogate function $S(.)$, which requires less computational expenses. Specifically, a large number of candidates will be drawn to evaluate the expected improvement at each iteration. The surrogate function approximates the contrastive loss over the set of drawn points, resulting in cheaper computational cost. Mathematically, the optimization problem is formulated as
$$\bm{\theta^*} = \arg \min_{\bm{\theta} \in \Theta} \mathcal{L}_{\bm{\theta}} (\bm{x}, \bm{x}^t)$$
\begin{algorithm}[htb]
  \caption{Sequential Model-Based Algorithm \cite{bergstra2011algorithms}}
  \label{TPE}
  Given $(\mathcal{L}_{\bm{\theta}^*}, M_0, I, S)$:
  \begin{enumerate}
    \item
    \textbf{Initialize} history $\mathcal{H} \leftarrow \emptyset$
    \item
    \textbf{For} iteration $i = 1$ \textbf{to} $I$:
    \begin{itemize}
      \item
        $\bm{\theta}^{*} \leftarrow \arg \min_{\bm{\theta}} S(\bm{\theta}, M_{i-1})$
      \item
      Evaluate $\mathcal{L}_{\bm{\theta}^*} (\bm{x}, \bm{x}^t)$
      \item
      Update $\mathcal{H} \leftarrow \mathcal{H}\cup ( \bm{\theta}^*,\mathcal{L}_{\bm{\theta}^*}(\bm{x}, \bm{x}^t))$
      \item
      Fit $M_i$ to $\mathcal{H}$
    \end{itemize}
  \item \textbf{Return} $\mathcal{H}$
  \end{enumerate}
\end{algorithm}
The SMBO algorithm is summarized in Algorithm ~\ref{TPE}, which attempts to optimize the Expected Improvement (EI) criterion \cite{bergstra2011algorithms}. Given a threshold value $t^*$, EI is the expectation under an arbitrary model $M$, that $\mathcal{L}_{\bm{\theta}} (\bm{x}, \bm{x}^t)$ will exceed $t^*$. Mathematically, we have:
\begin{equation}\label{EI}
    \text{EI}_{t^*} := \int_{-\infty}^{\infty} \max (t^* -t, 0)p_M(t|\bm{\theta})dt.
\end{equation}
While Gaussian-process approach models $p(t|\bm{\theta})$, the tree-structured Parzen estimator (TPE) models $p(\bm{\theta}|t)$ and $p(t)$, the it decomposes $p(\bm{\theta}|t)$ to two densitiy functions:
\begin{equation}\label{pdf}
    p(\bm{\theta}|t) = \begin{cases} l(\bm{\theta}) &\mbox{if } t < t^* \\
g(\bm{\theta}) & \mbox{if } t \geq t^* \end{cases} ,
\end{equation}
where $l(\bm{\theta})$ is the density function of candidate architectures corresponding to $\{\bm{\theta}^{(i)}\}$, such that  $\mathcal{L}_{\bm{\theta}^{(i)}} (\bm{x}, \bm{x}^t) < t^*$ and $g(\bm{\theta})$ is formed by the remaining architectures. TPE leverages multiple observations in the search space of NAS under non-parametric densities, enabling a learning process that derives multiple densities over the search space simultaneously. Besides, other difference between TPE and Gaussian-based approach is the selection of $t^{*}$. The TPE algorithm favours $t^{*}$ larger than the best observation of neural candidates, and then utilizes several points to construct the density $l(\bm{\theta})$, while Gaussian process favours more aggresive $t^{*}$ less than the best observation in the history $\mathcal{H}$. Thus, TPE can choose $t^{*}$ corresponding to some quantile of $t$, such that $p(t < t^{*}) = \gamma$.  As a result, the EI in Equation~\ref{EI} is reformed as:
\begin{multline}\label{EI_t}
    \text{EI}_{t^*}(\bm{\theta}) =  \int_{-\infty}^{t^*}(t^*-t)p(t|\bm{\theta})dt \\= \int_{-\infty}^{t^*}(t^*-t) \frac{p(\bm{\theta}|t)p(t)}{p(\bm{\theta})}dt \propto \bigg( \gamma + \frac{g(\bm{\theta})}{l(\bm{\theta})} (1-\gamma) \bigg)^{-1},
\end{multline}
where $\gamma$ denotes $p(t <t^*).$  The tree structure in TPE allows us to draw multiple candidates according to $l(.)$ and then evaluate them based on $g(\bm{\theta})/l(\bm{\theta})$. The TPE employed for the search strategy of NAS involves discrete-valued valuables, which represent the operations within neural cell's structure. The estimator samples a model for the search space by adaptively replacing the density in the vicinity of $K$ observations $\{\bm{\theta}^{(i)} \}_{1}^{K}$. The TPE treats the prior distribution of discrete variables as a vector of probability $p_i$, which has the same length as neural architecture's genotype vector. As a result, the posterior vector is proportional to $L p_i+C_i$, where $L$ is vector length of the neural genotype and $C_i$ is the counts of occurrences of choice $i$ in $\{\bm{\theta}^{(i)} \}_{1}^{K}$. Finally, the search time of each iteration of TPE can be scale linearly in $|\mathcal{H}|$ and the genotype vector length with sorted query of observation in $\mathcal{H}$.

\section{Experimental Results on CIFAR-10 and ImageNet} \label{experiment}
Our experiments on each dataset include two phases, neural architecture search (Sect~\ref{arcsearch}) and architecture evaluation (Sect~\ref{arceval}). It is worth mentioning that NAS algorithms have different strategy for selecting dataset for the search phase, while the same validation set is used in the evaluation phase. Pioneers such as NASNet and AmoebaNet assume the data constrain, which require the same dataset in search and evaluation. Although achieving remarkable results, the search cost of those approaches is extensively expensive due to the massive size of ImageNet ($\approx 14$ million samples). Following NAS algorithms relax the constrain, allowing search on smaller proxy dataset. The most popular proxy for ImageNet is attributed to CIFAR-10, which is widely used in later NAS algorithm. Regarding our work, we would like to preserve the data constrain since it is reasonable to assume that the neural solution found under the constrain may have higher performance.  In the search phase, we used only $10\%$ of unlabeled data ($5,000$ samples from CIFAR-10 \cite{krizhevsky2009learning} and approximately $12,000$ samples from ImageNet \cite{deng2009imagenet}) to search for models having the lowest contrastive loss mentioned in Equation ~\ref{final_loss} by CSNAS. The best architecture is scaled to a larger architecture in the validation phase, then trained from scratch on the train set and evaluated on a separate test set.

\subsection{Architecture Search for Convolution Cells}\label{arcsearch}

We initialize our search space by the operation generating set $\mathcal{O}$ as in Sect~\ref{NAconstruction}, which has been obtained by the most frequently chosen operators in \cite{zoph2018learning, real2019regularized,liu2018darts,liu2018progressive}. Each convolutional cell includes two inputs $C_{k-2}$ and $C_{k-1}$ (feature maps of two previous layers), a single concatenated output $C_{k}$ and $N$ intermediate nodes. 

\begin{table*}[h]
  \centering
  \caption{The performance (in term of test error) of state-of-the-art NAS algorithms on CIFAR-10. The search cost includes only searching time by SMBO-TPE algorithm, excluding the final architecture evaluation cost. \color{SeaGreen}$(\blacksquare)$ \color{black} Experiments use the same hyper-parameters setting reported in Appendix~\ref{appendix:arceval}. \color[HTML]{D3D3D3}$(\blacksquare)$ \color{black} Experiments with unknown/different evaluation settings.}
  \label{CNAS_CIFAR}
  \scalebox{1}{
  \begin{tabular}{cccccc}
    \toprule
    \textbf{Neural Architecture} & 
    \thead{\textbf{Test Error} \\ \textbf{(\%)}} & 
    \thead{\textbf{Params} \\ \textbf{(M)}} & 
    \thead{\textbf{Search Cost}\\ \textbf{(GPU days)}} & 
    \thead{\textbf{\#}\\\textbf{Ops}} & 
    \thead{\textbf{Search} \\ \textbf{Strategy}}\\
    \midrule
    DenseNet-BC \cite{huang2017densely} & $3.46$ & $25.6$  & - & -& Manual\\
    VGG11B ($2 \times$) \cite{nokland2019training} & $3.60$ & $42.0$  & - & -& Manual\\
    ResNet-1001 \cite{he2016identity} & $4.62$ & $10.2$  & - & -& Manual\\
    \midrule
    \rowcolor{SpringGreen} AmoebaNet-A + cutout \cite{real2019regularized} & $3.12$ & $3.1$ & $3150$ & $19$ & Evolution \\
    \rowcolor{SpringGreen} AmoebaNet-B + cutout \cite{real2019regularized} & $2.55 \pm 0.05$ & $2.8$ & $3150$ & $19$ & Evolution \\
    \rowcolor[HTML]{D3D3D3} CARS-I \cite{yang2020cars} & $2.62$ & $3.6$ & $0.4$ & $7$ & Evolution \\
    \rowcolor[HTML]{D3D3D3} Hierarchical evolution \cite{real2019regularized} & $3.75 \pm 0.12$ & $15.7$ & $300$ & $6$ & Evolution \\
    \rowcolor[HTML]{D3D3D3} LEMONADE \cite{elsken2018efficient} & $3.05$ & $4.7$ & $80$ & - & Evolution\\
    \rowcolor[HTML]{D3D3D3} NSGANet \cite{lu2019nsga} & $3.85$ & $3.3$ & $8$ & $7$ & Evolution\\
    \rowcolor[HTML]{D3D3D3} BlockQNN \cite{zhong2018practical} & $3.54$ & $39.8$ & $96$ & $8$ & RL \\
    \rowcolor{SpringGreen} ENAS \cite{pham2018efficient} + cutout & $2.89$ & $4.6$ & $0.5$ & $6$ & RL \\
    \rowcolor{SpringGreen} NASNet-A \cite{zoph2018learning} + cutout & $2.65$ & $3.3$ & $2000$ & $13$ & RL \\
    \rowcolor[HTML]{D3D3D3} BayesNAS \cite{zhou2019bayesnas} + cutout & $2.81 \pm 0.04$ & $3.4$ & $0.2$ & - & Gradient-based \\
    \rowcolor{SpringGreen} DARTS ($1^{st}$ order) \cite{liu2018darts} + cutout & $3.00 \pm 0.14$ & $3.3$ & $1.5$ & $7$ & Gradient-based \\
    \rowcolor{SpringGreen} DARTS ($2^{nd}$ order) \cite{liu2018darts} + cutout & $2.76 \pm 0.09$ & $3.3$ & $4$ & $7$ & Gradient-based \\
    \rowcolor{SpringGreen} GDAS\cite{dong2019searching} + cutout & $2.93$ & $3.4$ & $0.21$ & $7$ & Gradient-based\\
    \rowcolor[HTML]{D3D3D3} P-DARTS \cite{chen2019progressive} + cutout & $2.50$ & $3.4$ & $0.2$ & $7$ & Gradient-based \\
    \rowcolor[HTML]{D3D3D3} PC-DARTS \cite{xu2019pc} + cutout & $2.57 \pm 0.07$ & $3.6$ & $0.1$ & $7$ & Gradient-based \\
    \rowcolor[HTML]{D3D3D3} ProxylessNAS \cite{cai2018proxylessnas} +cutout & $2.08$ & $5.7$ & $4.0$ & - & Gradient-based\\
    \rowcolor{SpringGreen} MiLeNAS \cite{he2020milenas}  & $2.51 \pm 0.11$ & $3.87$ & $0.3$ & - & Gradient-based \\
    \rowcolor{SpringGreen} SNAS (moderate) \cite{xie2018snas} + cutout & $2.85 \pm 0.02$ & $2.8$ & $1.5$ & $7$ & Gradient-based \\
    \rowcolor{SpringGreen} GP-NAS \cite{li2020gp}  & $3.79$ & $3.90$ & $0.9$ & $7$ & Gaussian-Process-based\\
    \rowcolor{SpringGreen} PNAS \cite{liu2018progressive} & $3.41 +\pm 0.09$ & $3.2$ & $225$ & $8$ & SMBO\\

    \midrule
    \rowcolor[HTML]{D3D3D3} SSNAS\cite{kaplan2020self} & $2.61$ & - & $0.21$ & - & Gradient-based\\

    \midrule
    \rowcolor{SpringGreen} $\text{CSNAS}_{N=4}$ \textit{(ours)} + cutout $^{\dag}$ $\ddagger$ & $2.71 \pm 0.11$ & $3.5$ &  $0.25$& 7& SMBO-TPE\\
    \rowcolor{SpringGreen} $\text{CSNAS}_{N=5}$ \textit{(ours)} + cutout $^{\dag}$ $\ddagger$ & $2.66 \pm 0.07$ & $3.4$ &  $1$& 7& SMBO-TPE\\
    \bottomrule
    \\
    $\dag$ Results based on 10 independent runs.
    \end{tabular}}
\end{table*}

\begin{table*}[h]
  \centering
  \caption{The performance (in term of test error) of state-of-the-art NAS algorithms on ImageNet. \color{SeaGreen}$(\blacksquare)$ \color{black} Experiments use the same hyper-parameters setting reported in Appendix~\ref{appendix:arceval}. \color{Salmon}$(\blacksquare)$ \color{black} Experiments use the same hyper-parameters setting reported in Appendix~\ref{appendix:unnas-csnas}. \color[HTML]{D3D3D3}$(\blacksquare)$ \color{black} Experiments with unknown/different evaluation settings.} 
  \label{CNAS_IMG}
  \scalebox{1}{
  \begin{tabular}{ccccccc}
    \toprule
    \textbf{Neural Architecture} & 
    \thead{\textbf{Test Err. (\%)} \\ \textbf{Top-1 (Top-5)}} & 
    \thead{\textbf{Params} \\ \textbf{(M)}} & 
    \thead{\textbf{$\times +$}\\\textbf{(M)}} & 
    \thead{\textbf{Search Cost}\\ \textbf{(GPU days)}} & 
    
    \thead{\textbf{Search} \\ \textbf{Strategy}}\\
    \midrule
    Inception-v1 \cite{szegedy2015going}& $30.2 (10.1)$ &  $6.6$  & $1448$ & - & Manual\\
    Inception-v2 \cite{ioffe2015batch} & $25.2 (92.2)$ &  $11.2$  & $-$ & - & Manual\\
    MobileNet \cite{howard2017mobilenets}& $29.4 (10.5)$ &  $4.2$  & $569$ & - & Manual\\
    ShuffleNet (2$\times$)-v1 \cite{zhang2018shufflenet} & $26.4 (10.2)$ &  $\approx 5$  & $524$ & - & Manual\\
    ShuffleNet (2$\times$)-v2 \cite{ma2018shufflenet} & $25.1$ $(-)$ &  $\approx 5$  & $591$ & - & Manual\\
    \midrule
    \rowcolor{SpringGreen} NASNet-A \cite{zoph2018learning} $\ddagger$ & $26.0 (8.4)$ & $5.3$ & $564$ & $2000$ & RL \\
    \rowcolor{SpringGreen} NASNet-B \cite{zoph2018learning} $\ddagger$& $27.2 (8.7)$ $\ddagger$ & $5.3$ & $488$ & $2000$ & RL \\
    \rowcolor{SpringGreen} NASNet-C \cite{zoph2018learning} $\ddagger$& $27.5 (9.0)$ $\ddagger$ & $4.9$ & $558$ & $2000$ & RL \\
    \rowcolor{SpringGreen} AmoebaNet-A \cite{real2019regularized} $\ddagger$ & $25.5 (8.0)$ & $5.1$ & $555$ & $3150$ & Evolution \\
    \rowcolor{SpringGreen} AmoebaNet-B \cite{real2019regularized} $\ddagger$ & $26.0 (8.5)$ & $5.3$ & $555$ & $3150$ & Evolution \\
    \rowcolor{SpringGreen} AmoebaNet-C \cite{real2019regularized} $\ddagger$ & $24.3 (7.6)$ & $6.4$ & $570$ & $3150$ & Evolution \\
    \rowcolor{SpringGreen} AutoSlim \cite{yu2019autoslim} & $24.6 (-)$ & $8.3$ & $532$ & $224$ & Greedy\\
    \rowcolor[HTML]{D3D3D3} AtomNAS-A \cite{mei2019atomnas} & $25.4(7.9)$ & $3.9$ & $258$ & - & Dynamic network shrinkage\\
    \rowcolor[HTML]{D3D3D3} AutoNL-S \cite{li2020neural} & $22.3(6.3)$ & $5.6$ & $353$ & $32$ & Single-path\\
    \rowcolor[HTML]{D3D3D3} Single-Path \cite{stamoulis2019single} & $25.0(7.79)$ & - & - & $0.14$ & Single-path\\
    \rowcolor{SpringGreen} MnasNet-92\cite{tan2019mnasnet} & $25.2 (8.0)$ & $4.4$ & $388$ & - & RL\\
    \rowcolor{SpringGreen} PNAS \cite{liu2018progressive} $\ddagger$ & $25.8 (8.1)$ & $5.1$ & $588$ & $255$ & SMBO \\
    \rowcolor[HTML]{D3D3D3} PARSEC \cite{casale2019probabilistic}& $26.0(8.4)$ & $5.6$ & $548$ & $1$ & SMBO \\
    \rowcolor[HTML]{D3D3D3} FBNet-C \cite{wu2019fbnet} & $25.1 (-)$ & $5.5$ & $375$ & $20$ & SMBO \\
    \rowcolor[HTML]{D3D3D3} P-DART \cite{chen2019progressive}& $24.1 (7.3)$ & $5.4$ & $5.97$ & $2.0$ & Gradient-based\\
    \rowcolor[HTML]{D3D3D3} PC-DART \cite{xu2019pc}& $24.2 (7.3)$ & $5.3$ & $597$ & $3.8$ & Gradient-based\\
    \rowcolor[HTML]{D3D3D3} ProxylessNAS \cite{cai2018proxylessnas} & $24.9 (7.5)$ & $7.1$ & $465$ & $8.3$ & Gradient-based\\
    \rowcolor{SpringGreen} MiLeNAS\cite{he2020milenas} & $24.7 (7.6)$ & $5.3$ & $584$ & 0.3 & Gradient-based\\
    \rowcolor{SpringGreen} DARTS ($2^{nd}$ order) \cite{liu2018darts} $\ddagger$& $26.7 (8.7)$ & $4.7$ & $574$ & $4$ & Gradient-based \\
    \rowcolor{SpringGreen} SNAS (mild constraint) \cite{xie2018snas} & $27.3 (8.7)$ & $4.3$ & $522$ & $1.5$ & Gradient-based\\
    \rowcolor[HTML]{D3D3D3} RCNet \cite{xiong2019resource} & $27.8 (9.0)$ & $3.4$ & $294$ & $8$ & Gradient-based \\
    \rowcolor{SpringGreen} GDAS \cite{dong2019searching} & $26.0 (8.5)$ & $5.3$ & $581$ & $0.8$ & Gradient-based \\
    \midrule
    \rowcolor[HTML]{D3D3D3} SSNAS\cite{kaplan2020self} & $27.75 (9.55)$ & - & -& - & Gradient-based\\
    \rowcolor{Salmon} UnNAS-DARTS\cite{liu2020labels} + Jigsaw + cutout $\mathsection$ & $24.1 \pm 0.15 (-)$ & $5.2$ & $567$ & $2$ & Gradient-based\\

    \midrule
    \rowcolor{SpringGreen} $\text{CSNAS}_{N=5}$ \textit{(ours)} + cutout $\dag$  & $25.8 \pm 0.17(8.3)$ & $5.1$ & $590$  & $2.5$ & SMBO-TPE\\
    
    \rowcolor{Salmon} $\text{CSNAS}_{N=5}$ \textit{(ours)} + cutout $\mathsection$  & \textbf{$23.9 \pm 0.14(8.1)$} & $5.1$ & $590$  & $2.5$ & SMBO-TPE\\
    \bottomrule
    \\
    $\dag$ Results based on $10$ independent runs.\\
    $\mathsection$ Results based on $3$ independent runs.
    
    \end{tabular}}
\end{table*}

We create two searching spaces for CIFAR-10, denoted as $\text{CSNAS}_{N=4}$ and $\text{CSNAS}_{N=5}$, which are corresponding to the number of intermediate nodes $N$. With $N=4$, a configurable vector $\bm{\theta}$ representing a model have the length of 28 ($\bm{\theta}_{\text{normal}}$ = $\bm{\theta}_{\text{reduced}} = 14$), resulting in a search space of size $(8^{14})^2 \approx 2 \times 10^{25}$. We expand our search space by adding a single intermediate node, in the hope of finding a better architecture. $N=5$ is corresponding to configurable vector $\bm{\theta}$ of length 40, which tremendously surges the total number of possible architectures to $10^{36}$. Experiments involving ImageNet only use the latter search space with $N =5$.

We configure our C-SSL by two augmented views ($M=2$) for each sample with methods mentioned in Sect. \ref{SSL}, producing $2 \times (K-1)$ negative examples for each data instance in a minibatch of size $K$. The other two hyper-parameters $\tau$ and $\lambda$ in Equation~\ref{nce} and Equation~\ref{closs} are taken from the best experiment in \cite{misra2020self}, where $\tau = 0.07$ and $\lambda = 0.5$. Besides, MLPs $g(.)$ and $l(.)$ project encoded convolutional maps to a vector of size $p = 128$. Although we expect a minor impact of the above hyper-parameters, we will leave this tuning problem for further study.

We initialize the same prior density for each component of $\bm{\theta}$, which is that all operations have the same chance to be picked up at a random trial. $20$ random samplings start TPE, then $20K$ sample points are suggested to compute the expected improvement in each subsequent trial. We select only $20\%$ of best-sampled points having the greatest expected improvement to estimate next $\bm{\theta}$. We also observed that the number of starting trials is insensitive to the searching results while increasing the number of sampling points for computing expected improvement and lowering their chosen percentage ameliorate the searching performance (lower the overall contrastive loss).

We summarize the parameter settings and discovered cell architecture for all experiments in Sect~\ref{appendix:arcsearch}.

\subsection{Effectiveness Evaluation}\label{arceval}
We select the architecture having the best score from the searching phase and scale it for the validation phase. Within this paper's scope, we only scale the searched architecture to the same size as baseline models in the literature ($\approx 3M$). All weights learned from the searching phase had been discarded before the validation phase, where the chosen architecture is trained from scratch with random weights. 

Before analyzing the experimental results of CSNAS, we would like to outline several evaluation metrics for a NAS algorithm briefly. To begin with, we emphasize that predictive performance is a sufficient condition for good NAS algorithms. However, it is mandatory further to consider the versatility of NAS algorithms in different scenario. As mentioned in Section~\ref{merits}, supervised NAS algorithms likely result in better neural architectures than self-supervised NAS since they leverage full knowledge of training data (with annotations). On the other hand, self-supervised NAS offers us the opportunity to utilize additional unlabeled out-of-training samples, potentially lifting the curse of data when it comes to a scarcity scenario. Another evaluation metric for NAS algorithms is based on their ability to implement with limited computational resources. Finally, reported results from NAS algorithms are sensitive to the hyper-parameters setting of \textit{evaluation phase}, which may be attributed to the gain in overall performance. For example, DART used the same hyper-parameter setting (or reproduce other NAS with the same setting) with \cite{pham2018efficient,zoph2018learning,liu2018progressive,real2019regularized}; thus, the gain in accuracy can be entirely attributed to the effectiveness of search strategies. On the other hand, P-DART and PC-DART used a slightly higher regularization (increment of $0.1$ drop path probability) and larger batch size while remained the same learning rate as DART. The overall improvement may be slightly gained by such random effects in the evaluation phase. However, it cannot be deniable that P-DART and PC-DART offer us very highly efficient gradient-based NAS algorithms, tremendously reducing the computational expense in the search phase.

In the first block of Table~\ref{CNAS_CIFAR}, we compare our search with handcrafted architectures. Searched $\text{CSNAS}$ gains approximately $2\%$ and in test accuracy when comparing to ResNet-1001, while smaller gap (about $1\%$) on comparison to DenseNet-BC and VGG11B. We consider the results from manual designs a baseline for evaluating NAS algorithms since NAS's motivation is to search for better neural solutions automatically.

 In the second block, we report the results from SOTA supervised NAS algorithms. We perform the evaluation phase for CSNAS based on the exact setting used in \cite{liu2018darts,liu2018progressive,real2019regularized,zoph2016neural,pham2018efficient} to draw fair comparison with these supervised NAS algorithms. Regarding CIFAR-10, we report the performance of architecture searched by CSNAS in Table ~\ref{CNAS_CIFAR}. It is highlighted that $\text{CSNAS}_{N=4}$ achieved a slightly better result than DARTS with $16\times$ faster ($0.25$ in comparison to $4$), even though these algorithms share the same search space complexity. Moreover, $\text{CSNAS}_{N=5}$ can reach comparable results with  AmoebaNet and NASNet in a tremendously less computational expense ($1$ vs. $3150$ and $2000$, respectively). Similarly, the results of CSNAS on ImageNet are reported in Table ~\ref{CNAS_IMG}. Instead of transferring architecture from CIFAR-10, we directly search for the best architecture using 10\% of unlabeled samples from ImageNet (list of the images can be found in \cite{chen2020simple}. The performance of the network found by CSNAS appears to outperform DART and NASNET-A but to be slightly lower than AmoebaNet-C. Moreover, from both Table~\ref{CNAS_CIFAR} and ~\ref{CNAS_IMG}, the overall observation is that CSNAS possesses the ability to search for high-performance models, reaching comparable results with supervised NAS algorithms while using limited knowledge of data in search. 

The third block reports the performance of self-supervised NAS algorithms, including SSNAS and UnNAS-DART. It is noted that the two algorithms and our CSNAS search without using any data annotations. However, both SSNAS and UnNAS-DART used the whole training data (neglecting labels) for search, while we only used a small proportion of original training data. In the experiment on CIFAR-10, our CSNAS reaching slightly lower predictive performance than SSNAS. However, the hyper-parameter setting for evaluation phase is not reported within SSNAS's work, so it is hard to compare the performance of resulting models. Regarding ImageNet, our CSNAS obtains a gain of $~2\%$ in accuracy compared to SSNAS under the same evaluation setting. Since the evaluation of model derived by NAS algorithms is extremely sensitive to the evaluation setting, we adopt the same evaluation setting with UnNAS; which is reported in Section~\ref{appendix:unnas-csnas}, for a fair comparison. First, under the same setting, the model derived by CSNAS is slightly better than UnNAS-DARTS with solving jigsaw-puzzle SSL, achieving $23.8\pm 0.14$ top-1 accuracy in comparison to $24.1 \pm 0.15$ in UnNAS-DARTS. Second, it is worth noting that the search space of our CSNAS is much larger than self-supervised NAS competitors, which allows us to discover more potential neural candidates. Although searching on more massive search space, the time complexity of CSNAS is nearly the same as UnNAS due to the usage of surrogate models. In others word, the SMBO-TPE estimator enables us to approximate the expensive contrastive loss by cheaper surrogates. We report the hyper-parameter setting for evaluation phase in Section~\ref{appendix:arceval} and ~\ref{appendix:unnas-csnas}.

\subsection{Robustness of CSNAS to Loss of Information}\label{ablation}
\begin{figure}[t]
    \centering
    \includegraphics[width = 0.45\textwidth]{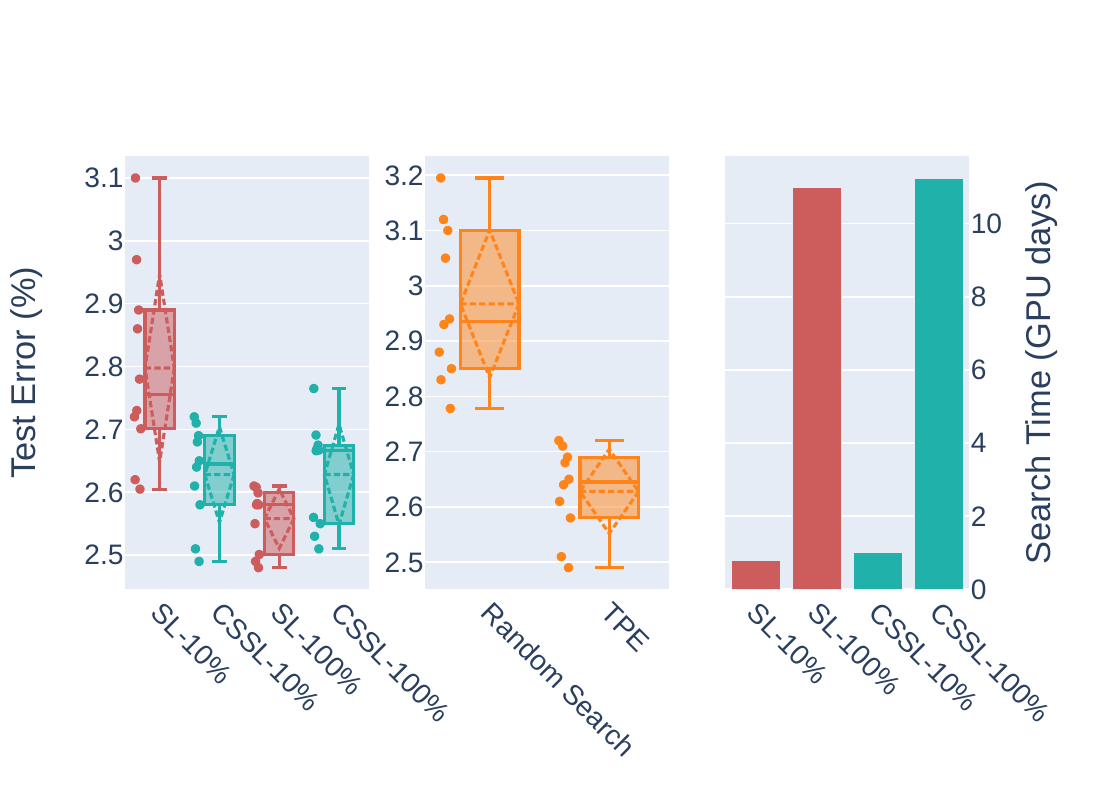}
    \caption{Experimental results on the robustness of CSNAS to the loss of information due to usage of C-SSL and surrogates. SL and CSSL are supervised and C-SSL. -$a\%$ represents the percentage of data used for the search phase.}
    \label{fig:ablation}
\end{figure}

In this section, we study the robustness of CSNAS to the loss of information, which is due to two reasons: (1) limited access to data's knowledge and (2) usage of surrogate models to approximate the true cost function. The first cause is that CSNAS leverages C-SSL, conducted on unlabelled and limited samples. Therefore, it should be outperformed by any other supervised NAS algorithms. Thus, we do not attempt to compare CSNAS with other supervised NAS algorithms. Instead, we would like to evaluate the loss of information caused by C-SSL on the overall performance of CSNAS. It is worth mentioning that such loss is tough to measure since it depends on the dataset of interest. In other words, the loss of information will be different when it comes to different datasets. The second cause is due to surrogate models, which possess a high semblance to the true expensive loss function. Although surrogates' usage helps reduce the computational expense for the search phrase, the approximation process may induce other losses of information caused by surrogates. Since universally evaluating such loss is impossible, we only evaluate the loss of information from CSNAS on the CIFAR-10 dataset. 

In the experimental design, we include two components: (1) C-SSL and (2) Usage of surrogate models, which both induce the loss of information. First, we evaluate such loss by fixing (2) and changing (1) to supervised search. As a result, we can approximate the loss of information caused by C-SSL on CIFAR-10. The left panel of Figure~\ref{fig:ablation} shows that the loss of information caused by C-SSL does not exist from experiments on $10\%$ of data samples without any annotation (CSSL-$10\%$ vs. SL-$10\%$ ) since the model searched by C-SSL outperforms that derived by supervised learning. On the other hand, the loss of information appears when it comes to $100\%$ of the dataset (CSSL-$100\%$ vs. SL-$100\%$). As expected, the model found by supervised learning achieves a better result than that from C-SSL. However, we observe that such loss is minimal in terms of validation accuracy.
Moreover, it appears that the loss of information caused by the proportion of samples used is extremely small (CSSL-$10\%$ vs. CSSL-$100\%$) since the performance of derived architectures is nearly the same. The last detail of interest in this experiment is investigating the time complexity of the search phase due to C-SSL. As mentioned before, C-SSL requires multiple augmented views of a given input. From the right panel of Figure~\ref{fig:ablation}, we can see that the computational expense increased by C-SSL is extremely minimal. Hence, the additional search cost is inconsiderable.

Regarding the second component, we compare the result from SL-$100\%$ with other supervised NAS addressed in Table~\ref{CNAS_CIFAR}. It is noting that SL-$100\%$ is conducted by SMBO-TPE under supervised learning with full access to CIFAR-10, achieving $2.56\%$ in test error. Referring to Table~\ref{CNAS_CIFAR}, SMBO-TPE mostly outperforms other search strategies (except MiLeNAS) under the same evaluation setting, which emphasizes the effectiveness of SMBO-TPE in comparison to other optimization strategies. We also observe the effectiveness of the Tree Parzen Estimator in comparison to baseline random search, which is highlighted in the middle panel of Figure~\ref{fig:ablation}. Moreover, SMBO-TPE constructs a probability model of the cost function in Equation~\ref{closs} and then uses it to find the most potential neural candidates to evaluate the true function. Before evaluating the true cost function, we sample $20K$ genotypes that defined neural candidates from the search space and only selected the top $20\%$ of this population. The optimal neural architecture is founded by approximately $500$ rounds of evaluating the true cost function. By periodically computing the true cost, the loss of information caused by surrogates is mitigated by providing a better approximation of the loss landscape.

In conclusion of this section, we observe no loss of information caused by CSSL on $10\%$ of the dataset. However, such loss appears on $100\%$ of samples, where supervised search outperforms C-SSL. Besides, the loss of information caused by using surrogates is nearly the same as other competitors. Finally, SMBO-TPE aggressively samples the most promising neural candidates for each exact evaluation of the true cost function, which offers well-approximation for the lost landscape. As a result, the loss of information caused by surrogates is alleviated.

\section{Case Study}\label{case_study}
This section investigates the effectiveness of CSNAS on a practical case study, which involves skin lesion classification. The under-investigated problem is an excellent example of a data scarcity scenario, where labeled data is costly (requires expert knowledge), and unlabelled data have zero-cost for annotations. Self-supervised NAS algorithms are suitable for such cases. Moreover, at the same time as our study, CSNAS is the second work considering skin lesion classification. \cite{kwasigroch2020neural} used network morphism to search neural architecture for skin lesion classification. However, they only consider binary classification problems while we performed multi-class classification. We are hoping to compare our CSNAS to other self-supervised NAS. Unfortunately, we cannot find the official implementation of SSNAS and UnNAS in the meantime. It is noted that reproducing the work without official implementation may induce inaccurate observation and false comparison \cite{li2020random,sciuto2019evaluating}. Therefore, we can only compare our CSNAS to the conventional approach for skin lesion classification, which is transfer learning.

We organize the section as follows: Section~\ref{isic} summarizes the classification problem with a detailed introduction of the ISIC-2019 database. Section~\ref{experiment} gives a detailed experimental setting for the case study. Finally, we report the quantitative results in Section~\ref{effectiveness}. To avoid confusion with searched architectures from CIFAR-10 and ImageNet, we named architecture searched on ISIC-2019 as DermoCSNAS.

\subsection{ISIC-2019 Dataset}\label{isic}
The dataset of interest is International Skin Imaging Collaboration database (ISIC 2019)\cite{tschandl2018ham10000,codella2018skin,combalia2019bcn20000}, which includes a public train set of 25,331 labeled images and a private test set of 8,238 unlabeled images. For the illustration of the proposed neural solution, the unlabeled samples in the test set are used for searching deep models in a self-supervised manner, and then the found architecture is conventionally trained on the train set to perform classification. It is highlighted that the public train set and private test set provided by ISIC 2019 are not overlapped. The ISIC 2019 public train set originally contains nine classes, which are melanoma (MEL), melanocytic nevus (NV), basal cell carcinoma (BCC), actinic keratosis (AK), benign keratosis (BKL), dermatofibroma (DF), vascular lesion (VASC), squamous cell carcinoma (SCC) and unknown disease (UNK). Since the number of unknown training samples is zero, we only consider defined disease, resulting in a multi-label classification of 8 classes. Table~\ref{isic2019} depicts the distribution of classes and search-train-test-validation splits (follows ratio $80\%-5\%-15\%$) for our experiment. Moreover, we also publish the list of train/test/validation samples in the GitHub repository for reproduction purposes.
\begin{table}[t]
  \centering
  \caption{ISIC-2019 dataset in detail. The data used for search is out-of-training and accounted for 32.5\% of training data.} 
  \label{isic2019}
  \scalebox{0.8}{
  \begin{tabular}{cccc}
    \toprule
    \textbf{Dataset} & 
    \textbf{Used Phase} & 
    \thead{\textbf{Number of} \\ \textbf{samples}} & 
    \textbf{Split}\\
    \midrule
    \midrule
    ISIC-2019 private test set & Search & $8,238$ & - \\
    \midrule
    \midrule
    ISIC-2019 public train set & Validation &  \makecell{ Train \\ Validation \\ Test } & \makecell{20,265\\ 1,290 \\ 3,776 }\\
    \midrule
     & & Total & 25,331\\
    \midrule
    \textbf{Skin Disease} & 
    \textbf{Annotation}&
     & 
    \textbf{Distribution}\\
    \midrule
    \midrule
    Melanoma & MEL & 4,522& 17.85\%\\
    Melanocytic  Nevus & NV & 12,875 & 50.83\% \\
    Basal Cell Carcinoma & BCC & 3,323 & 13.12\% \\
    Actinic Keratosis & AK & 867 & 3.42\% \\
    Benign Keratosis & BKL & 2,624 & 10.36\% \\
    Dermatofibroma & DF & 239 & 0.94\% \\
    Vascular Lesion & VASC & 253 & 1\% \\
    Squamous Cell Carcinoma & SCC & 628 & 2.48\% \\
    Unknown & UNK & 0 & 0\% \\
    \bottomrule
    \end{tabular}}
    
\end{table}
\subsection{Experimental Setup}\label{experiment}
Our experiment includes two phases, which are (1) searching neural architecture on unlabeled data (ISIC private test set) under a SSL manner and (2) evaluating discovered neural net on labeled samples (ISIC public train set). To preserve our procedure's robustness, we wholly removed the learned model weights after the searching phase. Then, we found that models was trained from scratch using a random initialization in the validation phase.
\subsubsection{Search Phase}
Our configuration for this experiment is similar to CIFAR-10 and ImageNet. First, we investigate two configurations for searching spaces, corresponding to $N=4$ and $N=5$ intermediate nodes. The architectures found under these settings are denoted as DermoCSNAS-4 and DermoCSNAS-5, in which the encoded vector $\bm{\theta}$ of a neural candidate DermoCSNAS-4 is a 28-dimensional vector, while the corresponding vector for DermoCSNAS-5 is a 40-dimensional vector. As a result, the number of possible neural candidates exponentially increases ($5\times 10^{10}$ times) when added to only one intermediate node. Second, we generate neural candidates using the operation generating set $\mathcal{O}$ mentioned in section-- and perform C-SSL with two augmented views $(M=2)$ for each sample, resulting in $2\times (K-1) = 298$ negative examples for each data point in a mini-batch of $K =150$ samples. Each candidate contains only $8$ layers with $16$ initial channels, which is trained using momentum SGD with the learning rate of $0.001$ and momentum of $0.9$. The hyper-parameter for noise contrastive estimator in Equation is set as $[\tau, \lambda] = [0.07, 0.5]$. Finally, we initialize the TPE by 20 random samplings, followed by $20K$ suggested points for computing the expected improvement of each trial. There only $20\%$ of best candidates having the largest expected improvement is mutated for estimating the next $\bm{\theta}$. 

\subsubsection{Validation Phase}\label{val_phase}
First, we construct the final model by expanding the depth (number of layers) and the width (number of initial channels) from the discovered neural cell in the search phase. In both configurations of $N=4$ and $N=5$, we stack $18$ layers of the founded cell with $48$ initial channels since we aim to provide a hardware-aware deep model, which is restricted under $600$ number of multiply-add operation. Noted that all of the reduction cells are in one-half and two-thirds of the depth of model, in which all chosen operations use a stride of 2. Second, we augmented training samples by downsampling to $256\times 256$ before random cropped into $224 \times 224$, then randomly apply horizontal and vertical flipping. Moreover, we prevent over-fitting by linearly increasing path drop out of $0.2$ as in\cite{zoph2016neural,zoph2018learning,liu2018darts,liu2018progressive,real2019regularized}; cutout of length $16$\cite{devries2017improved} and a small auxiliary classifier (at two-third of model's depth) with weight of $0.4$. Finally, the model is trained with a batch size of $128$ using SGD optimizer, which is initialized by $0.02$ learning rate and $0.9$ weights decay. The learning rate has a decay rate of $0.99$ for every $3.5$ epoch.

\subsection{Effectiveness analysis}\label{effectiveness}
\begin{table}[t]
  \centering
  \caption{Comparisons between state-of-the-art neural architectures and our proposed model, in terms of test accuracy and model complexity. The results of our $\text{DermoCSNAS}$ are mean and variance from 10 independent runs, which use the same set of training hyper-parameters in Section~\ref{val_phase}. Moreover, we use the same experimental setting (data augmentation and training parameters) across all transfer learning experiments in order to provide unbiased results.} 
  \label{cls_res}
  \scalebox{0.8}{
  \begin{tabular}{cccccc}
    \toprule
    \textbf{Neural Architecture} & 
    \thead{\textbf{Params (\%)} \\ \textbf{(M)}} & 
    \thead{\textbf{FLOPS} \\ \textbf{(G)}} & 
    \thead{\textbf{Test Acc.}\\\textbf{(\%)}} &
    \thead{\textbf{Melanoma}\\\textbf{F1-score}}\\
    \midrule
    DPN-131 \cite{chen2017dual} & $ 79.3$  &  $16.0$  & $86.23$ & $0.79$\\
    EfficientNet-B0 \cite{tan2019efficientnet}  & $ 5.3$  &  $0.39$  & $82.14$ & $0.80$\\
    ResNet152 \cite{he2016deep} & $ 60.0$  &  $11.3$  & $84.00$ & $0.75$\\
    ResNet101 \cite{he2016deep} & $ 44.6$  &  $8.0$  & $87.76$ & $0.81$\\
    Inception-v4 \cite{szegedy2015going} & $ 46.0$  &  $12.3$  & $85.99$ & $0.79$\\
    Inception-ResNet-v2 \cite{szegedy2017inception}& $ 55.8$  &  $11.75$  & $87.53$ & $0.80$\\
    NASNet \cite{zoph2018learning} & $ 88.9$  &  $24.0$  & $87.80$ & $0.82$\\
    PNASNet\cite{liu2018progressive} & $ 86.1$  &  $25.2$  & $87.87$ & $0.81$\\
    SENet101\cite{hu2018squeeze} & $ 49.2$  &  $8.0$  & $87.55$ & $0.81$\\
    SENet154\cite{hu2018squeeze} & $ 145.8$  &  $42.3$  & $88.00$ & $0.83$\\
    Xception \cite{chollet2017xception} & $ 23.0$  &  $8.4$  & $87.18$ & $0.80$\\
    \midrule
    $\text{DermoCSNAS}_{N=4}$  & 3.55 & 0.503 & $87.94 \pm 0.04$ & $0.83$\\
    $\text{DermoCSNAS}_{N=5}$  & $\bm{4.15}$ & $\bm{0.560}$ & $\bm{88.68} \pm 0.02$ & $\bm{0.84}$\\

    \bottomrule
    
    \end{tabular}}
\end{table}
We depict the effectiveness of our approach by comparing it with state-of-the-art models in Table~\ref{cls_res}. It is noted that other state-of-the-art architectures are fine-tuned with pre-trained weights (transfer learning), while our model is trained from scratch on the ISIC dataset. Hence it is inevitable that lower-level features in early layers of our model are learned from skin lesion images. In contrast, we need to accept inherited lower-level features from domain datasets (ImageNet or CIFAR-10) once performing transfer learning and fine-tuning pre-trained models. As discussed in the beginning of this section, we cannot reproduce other self-supervised NAS algorithms without significant inaccurate results. Thus, we instead evaluate the performance of DermoCSNAS by comparing to the most conventional approach of skin lesions classification, which is transfer learning. Within the scope of this study, we compare our discovered model with state-of-the art deep convolution networks, which include Efficient-B0\cite{tan2019efficientnet}, ResNes-101 and ResNet-152\cite{he2016deep}, Inception-v4\cite{szegedy2015going}, Inception-ResNet-v2\cite{szegedy2017inception}, DPN-131\cite{chen2017dual}, Xception\cite{chollet2017xception}, SENet-101 and SENet-154\cite{hu2018squeeze}, NASNet\cite{zoph2018learning}, PNASNet\cite{liu2018progressive}.


\begin{table}[t]
  \centering
  \caption{Classification report of our $\text{DermoCSNAS}_{N=5}$. The searched neural architecture possesses high predictive performance. Besides, the precision of Melanoma (cancerous class) is well-established.} 
  \label{clasrep}
  \scalebox{0.8}{
  \begin{tabular}{cccccc}
    \toprule
    \textbf{Classes} & 
    \textbf{Precision} & 
    \textbf{Recall} & 
    \textbf{F-1 score} &
    \textbf{Support}\\
    \midrule
    Melanoma & $0.84$  &  $0.81$  & $0.83$ & $647$\\
    Melanocytic Nevus & $0.92$  &  $0.95$  & $0.93$ & $1991$\\
    Basal cell carcinoma & $0.87$  &  $0.90$  & $0.88$ & $465$\\
    Actinic Keratosis  & $0.79$  &  $0.62$  & $0.69$ & $119$\\
    Benign Keratosis  & $0.76$  &  $0.76$  & $0.76$ & $381$\\
    Dermatofibroma  & $0.85$  &  $0.78$  & $0.81$ & $36$\\
    Vascular & $0.97$  &  $0.72$  & $0.82$ & $39$\\
    Squamous cell carcinoma & $0.75$  &  $0.67$  & $0.71$ & $98$\\
    \midrule
    
    Macro average & $0.84$ & $0.78$ & $0.81$ & 3776\\
    Weighted average & $0.88$ & $0.88$& $0.88$ \\
    
    \bottomrule
    
    \end{tabular}}
\end{table}

First, the searched neural architecture under the best setting $(\text{DermoCSNAS}_{N=5})$ outperforms other human-crafted model, gaining $0.68\%-4.68\%$ (in compare to SENet154 and ResNet152, respectively). We also observed no trade-off between the model complexity and the overall performance since our model possesses the smallest number of parameters and the number of multiply-add operations. 
Second, our model has the best F-1 score from malignant classes (Melanoma) at $0.84$, while others are ranging from $0.75$ to $0.83$. It is crucial since our neural intelligence makes no trade-off between the overall accuracy and the balance of precision and recall from cancerous class. The detailed classification report is showed in Table~\ref{clasrep}.

\section{Discussion}
\subsection{Implication}
Beyond the experimental results on mainstream datasets (CIFAR-10 and ImageNet) and the case study of skin lesion classification (ISIC-2019), we would like to discuss the general principles that can be taken from our CSNAS. The advantages of our CSNAS are mainly from the power of representation learning of C-SSL, which can effectively learn the underlying representation with access to a small proportion of training data without labels. The advantage benefits the generalization of large-scale implementation. In a data-abundant scenario, we can randomly draw a small proportion of training data and ignore annotations for the search phase. Our study on ImageNet used a benchmarking sub-1\% and 10\% dataset (commonly used in SSL and self-training results), while in ISIC-2019, we used additional unlabeled data for the search phase. It is worth mentioning that we often cope with computer vision problems involving data scenarios similar to our case study, in which data annotations are costly due to extensive human experts. Hence, CSNAS offers an efficient NAS algorithm in such scenarios.

\subsection{Threats to Validity}
Threats to internal validity include the consistency of reproducibility of work. The notorious challenge of NAS-related research is the reproducible ability \cite{li2020random,sciuto2019evaluating}. Early NAS algorithms require extensive computational resources to search for a model that is entirely unavailable for large-scale applications. Sub-sequence algorithms tackle the issue by search on a smaller configuration space and implement a more efficient search strategy. However, the results are not comparable to each other since the experimental setting is extremely sensitive to the final results. A partial solution for the issue can be delivered through the quantitative evaluation among pre-trained models. However, we still hope to know a clear insight into the comparison of NAS algorithms. Several attempts in NAS-related research, including \cite{ying2019bench,dong2020bench,zela2020bench,siems2020bench}, tackle the reproducible issue efficiently. However, their implementation only narrows in the mainstream dataset - CIFAR-10. 

Threats to external validity involve the generalization of our work on different domain-specific datasets and different computer vision tasks. First, our proposed CSNAS fundamentally is based on contrastive self-supervise learning, which possesses a strong image representation capability even when using a small proportion of training samples without annotations. Thus, CSNAS benefits neural architecture search on data scarcity scenario, where labeled data is costly. Our case study shows that DermoCSNAS achieves high predictive performance compared to the dominant competitor field - transfer learning.  Moreover, we keep the data constrain for the search phase and evaluation phase, in which found model is discovered in the same dataset as the evaluation phase. Intuitively, the constrain offers us a robust model on the domain-specific dataset. 
Regarding the transferability of searched models to different computer vision tasks, the improvement is consistent when we investigate semantic segmentation \cite{liu2020labels}, object detection \cite{chen2019detnas} and adversarial learning \cite{gong2019autogan}.

\section{Conclusion} \label{conclusion}
We have introduced CSNAS, an automated NAS algorithm that completely alleviates the expensive cost of data labeling. Furthermore, CSNAS performs searching on natural discrete search space of NAS problem via SMBO-TPE, enabling competitive/matching results with state-of-the-art algorithms.

There are many directions to conduct further study on CSNAS. For example, computer vision tasks, which involve medical images, are usually considered to lack training samples. This task requires substantially expensive data curation cost, including data gathering and labeling expertise. Another possible CSNAS improvement is investigating further baseline SSL methods, which potentially ameliorates current CSNAS benchmarks.

\section*{Acknowledgments}
Effort sponsored in part by United States Special Operations Command (USSOCOM), under Partnership Intermediary Agreement No. H92222-15-3-0001-01. The U.S. Government is authorized to reproduce and distribute reprints for Government purposes, notwithstanding any copyright notation thereon. \footnote{The views and conclusions contained herein are those of the authors and should not be interpreted as necessarily representing the official policies or endorsements, either expressed or implied, of the United States Special Operations Command.}


\bibliographystyle{unsrt}
\bibliography{output}

\appendix
\section{Experimental Details}
\subsection{Neural Architecture Search} \label{appendix:arcsearch}
\subsubsection{Experimental setting}
For CIFAR-10 dataset, we use $5000$ class-balanced images to search for the best architecture. While architecture search on ImageNet use the same list of samples as \cite{chen2020simple}.

We use a mini-batch of size $K=150$, resulting in $298$ negative samples corresponding to a single data point each mini-batch. We accelerate the searching time by using small architecture candidates, which include 8 layers and 32 channels. For optimizing the weights $w$ in memory-bank $\mathcal{M}$, we use momentum SGD with learning rate $\eta_{w} = 0.001$ and momentum $0.9$. We set $\tau = 0.07$ and $\lambda = 0.5$ for the noise contrastive estimator and contrastive loss as in \cite{misra2020self}. We set up the TPE sampler as in Sect~\ref{arcsearch} with zero initialization for $\bm{\theta} =(\bm{\theta}_{\text{normal}},\bm{\theta}_{\text{reduce}})$. 

\subsubsection{Neural Cell discovered by CSNAS}
\begin{figure}[h]
\begin{center}
\includegraphics[width=0.3\linewidth]{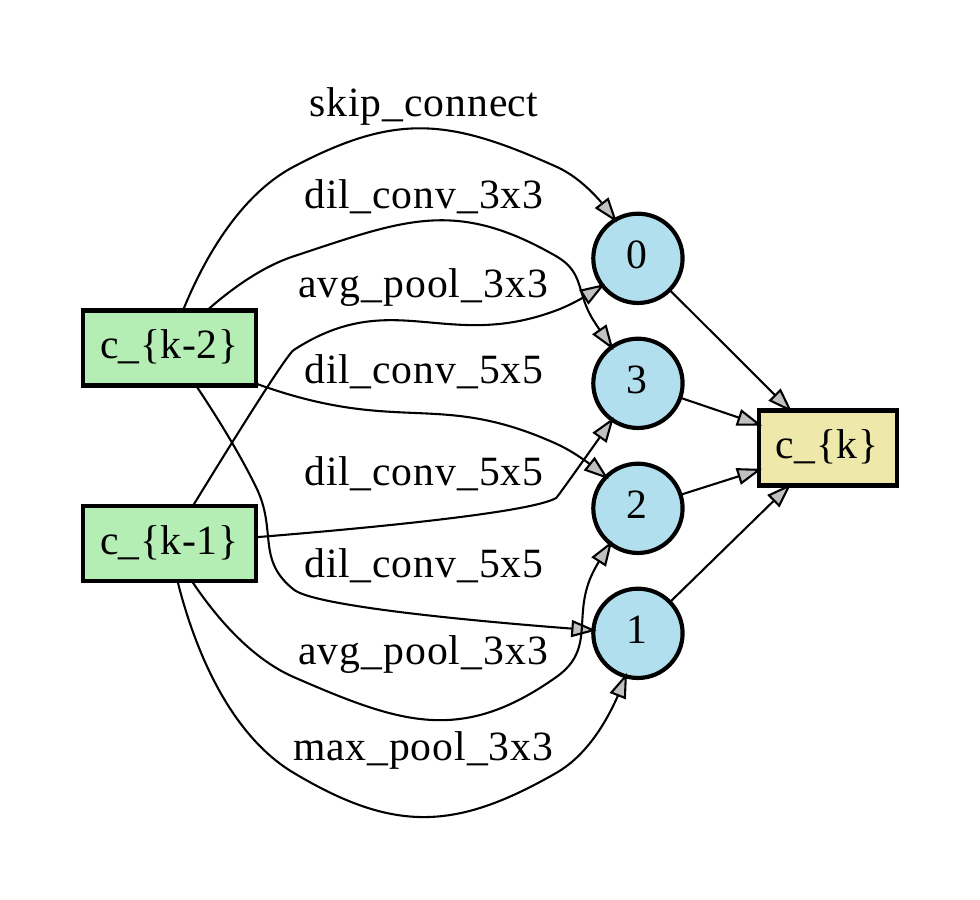}
\includegraphics[width=0.3\linewidth]{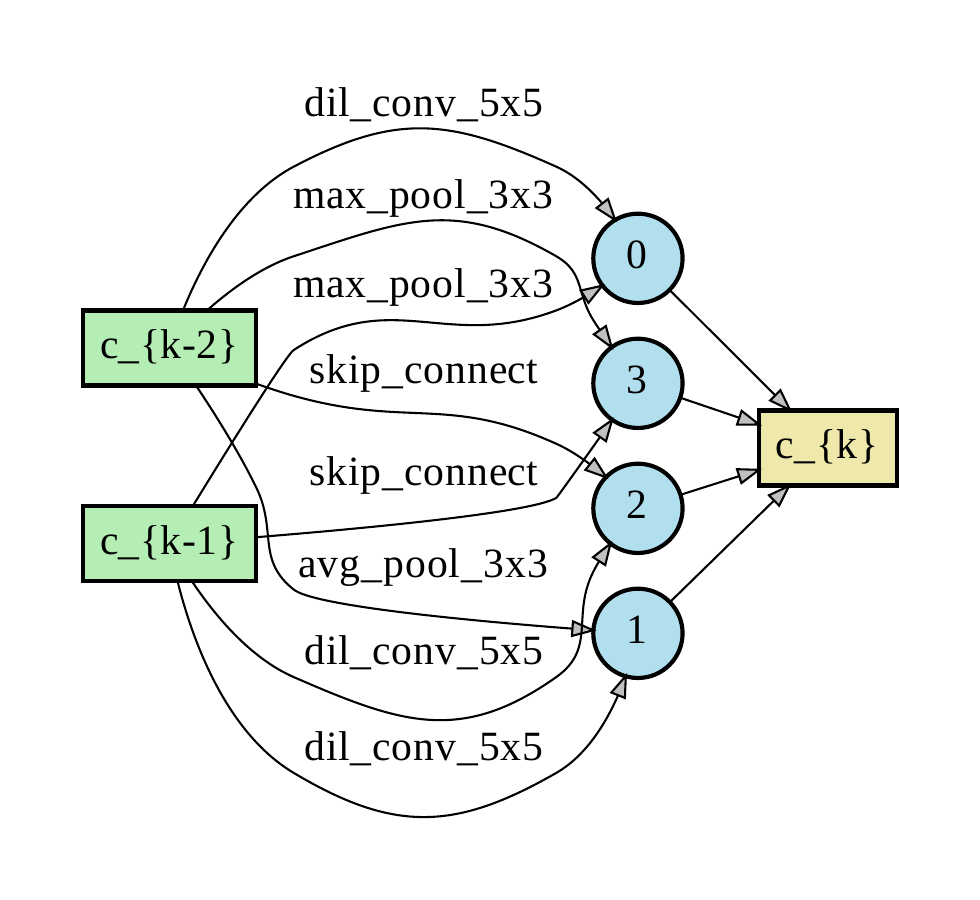}
\includegraphics[width=0.45\linewidth]{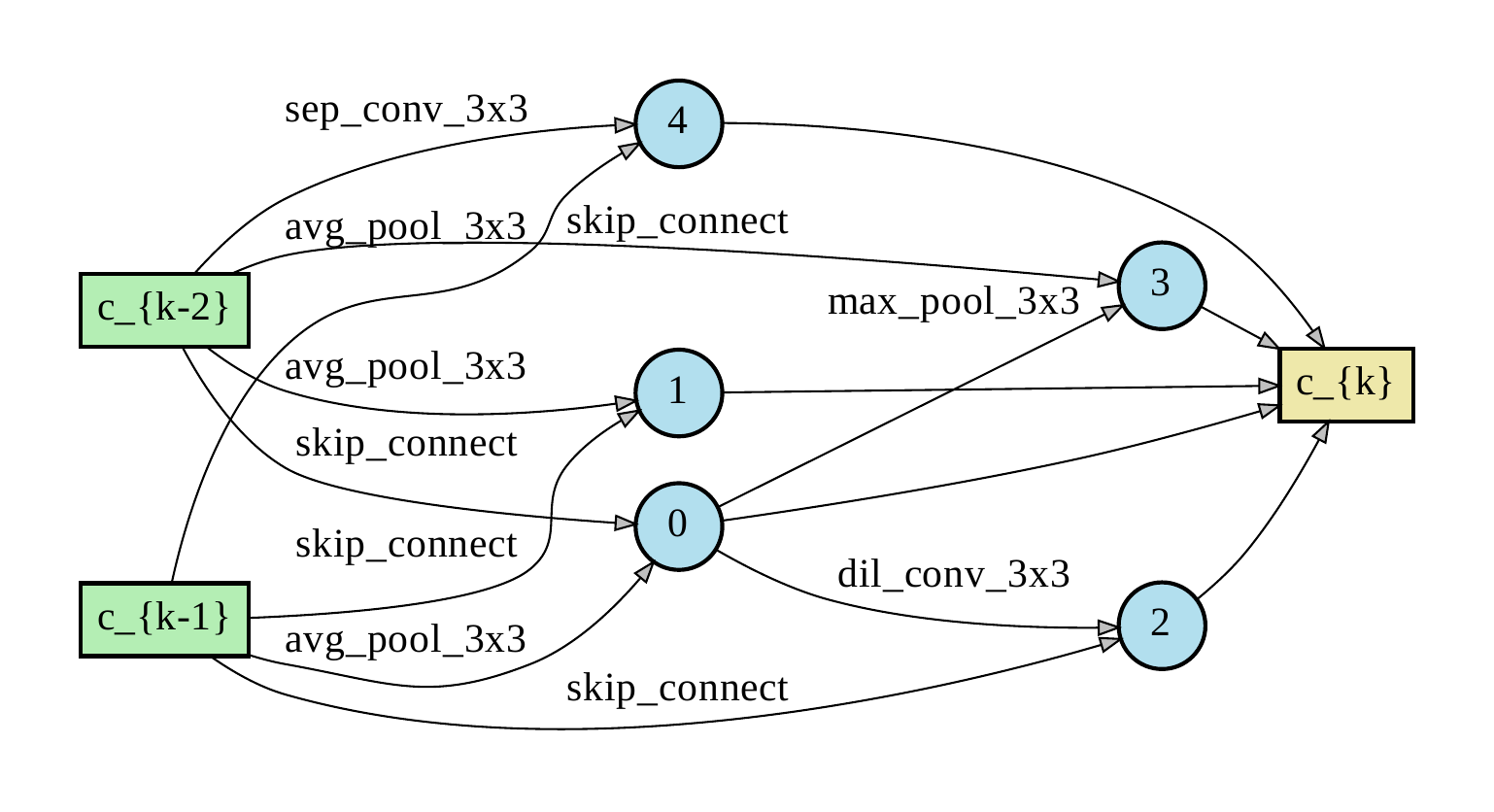}
\includegraphics[width=0.45\linewidth]{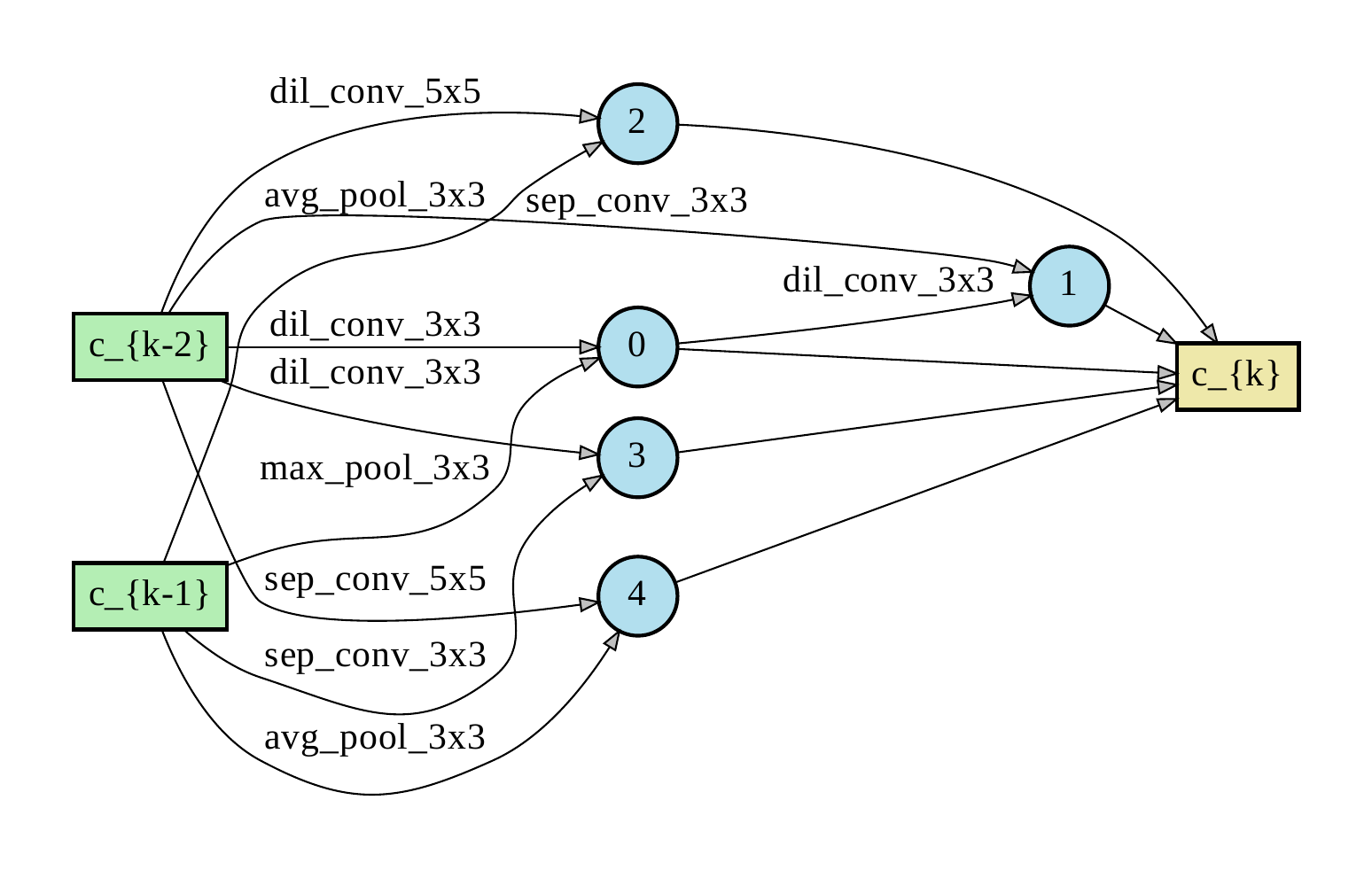}

\end{center}
\caption{Cell architecture of $\text{CSNAS}_{N=4}$ and $\text{CSNAS}_{N=5}$  searched on CIFAR-10.} 
\label{c1}
\end{figure}

\begin{figure}[h]
    \begin{center}
    \includegraphics[width=0.5\linewidth]{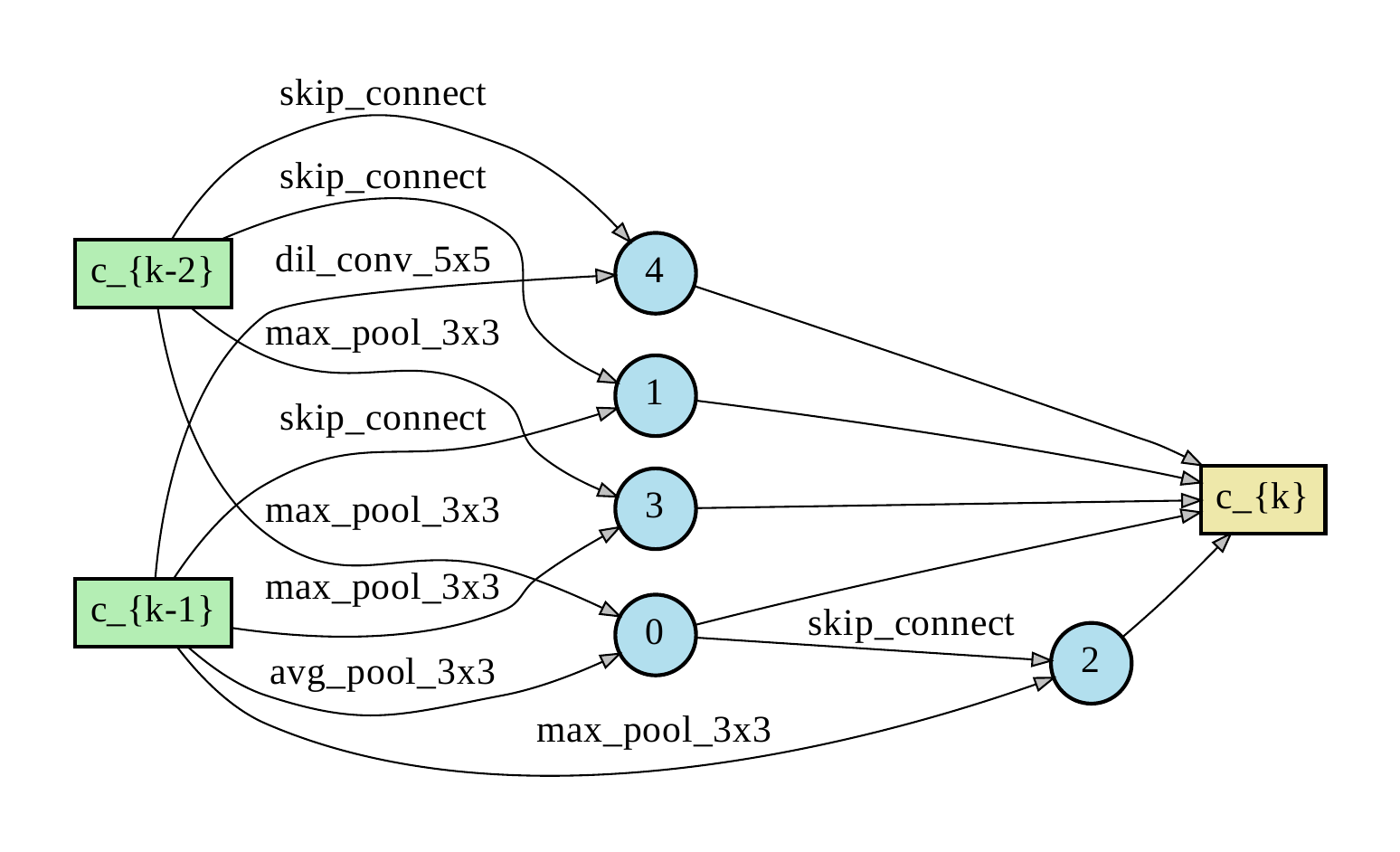}
    \includegraphics[width=0.33\linewidth]{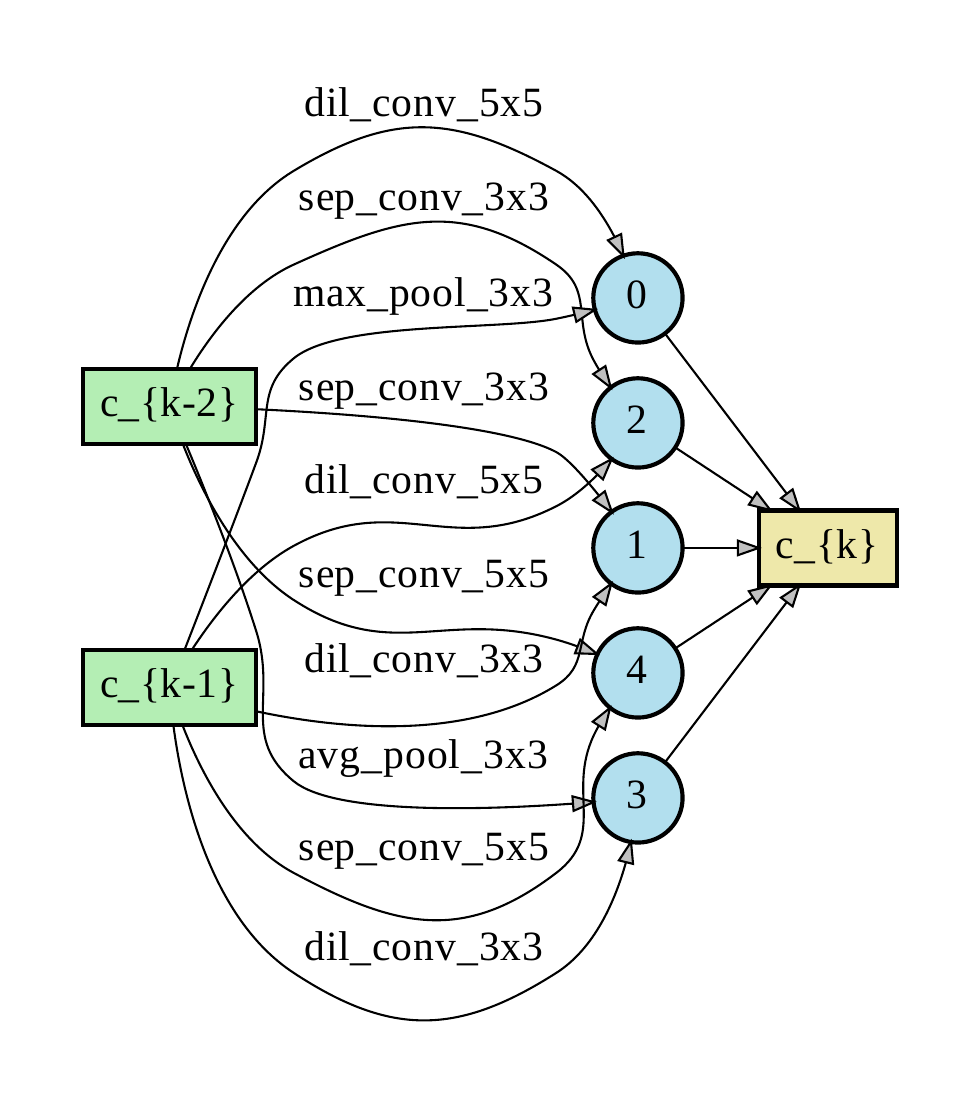}
    \end{center}
    \caption{Cell architecture of $\text{CSNAS}_{N=5}$ searched on ImageNet-1\%. The vector $\bm{\theta}$ representing each neural candidate is of length 20.}
    \label{c2}
\end{figure}

\begin{figure}[h]
    \centering
    \includegraphics[width = 0.2\textwidth]{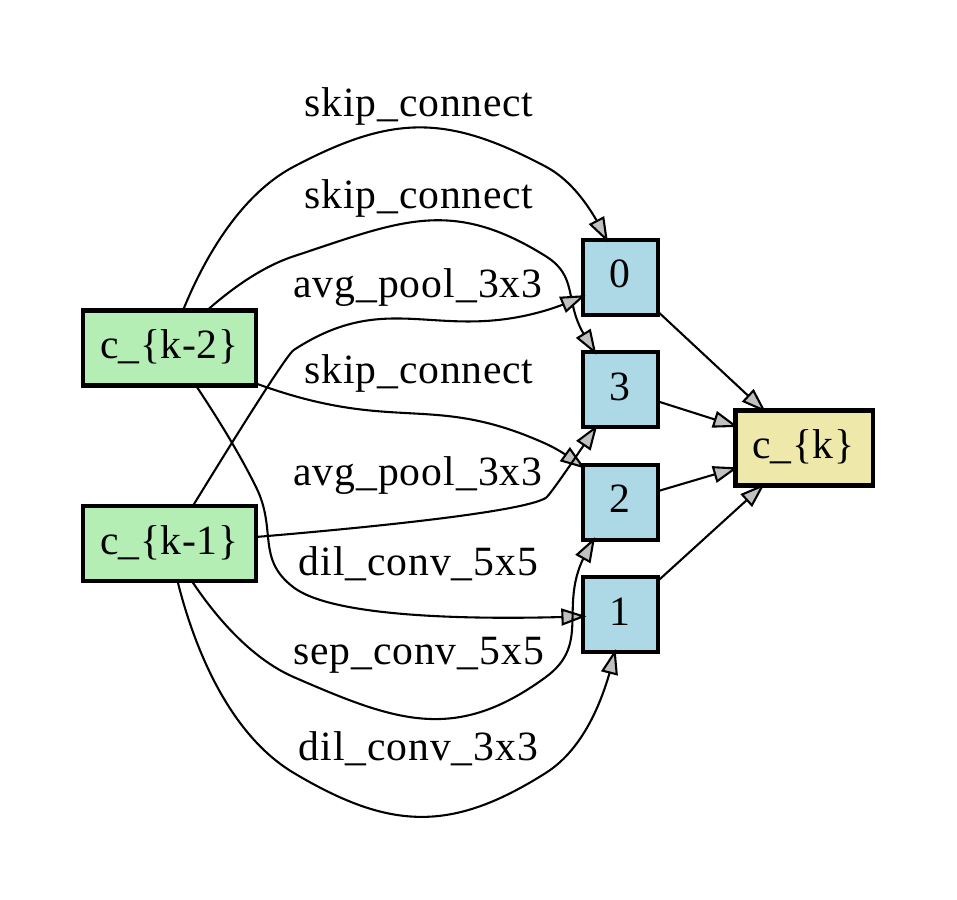}
    \includegraphics[width = 0.25\textwidth]{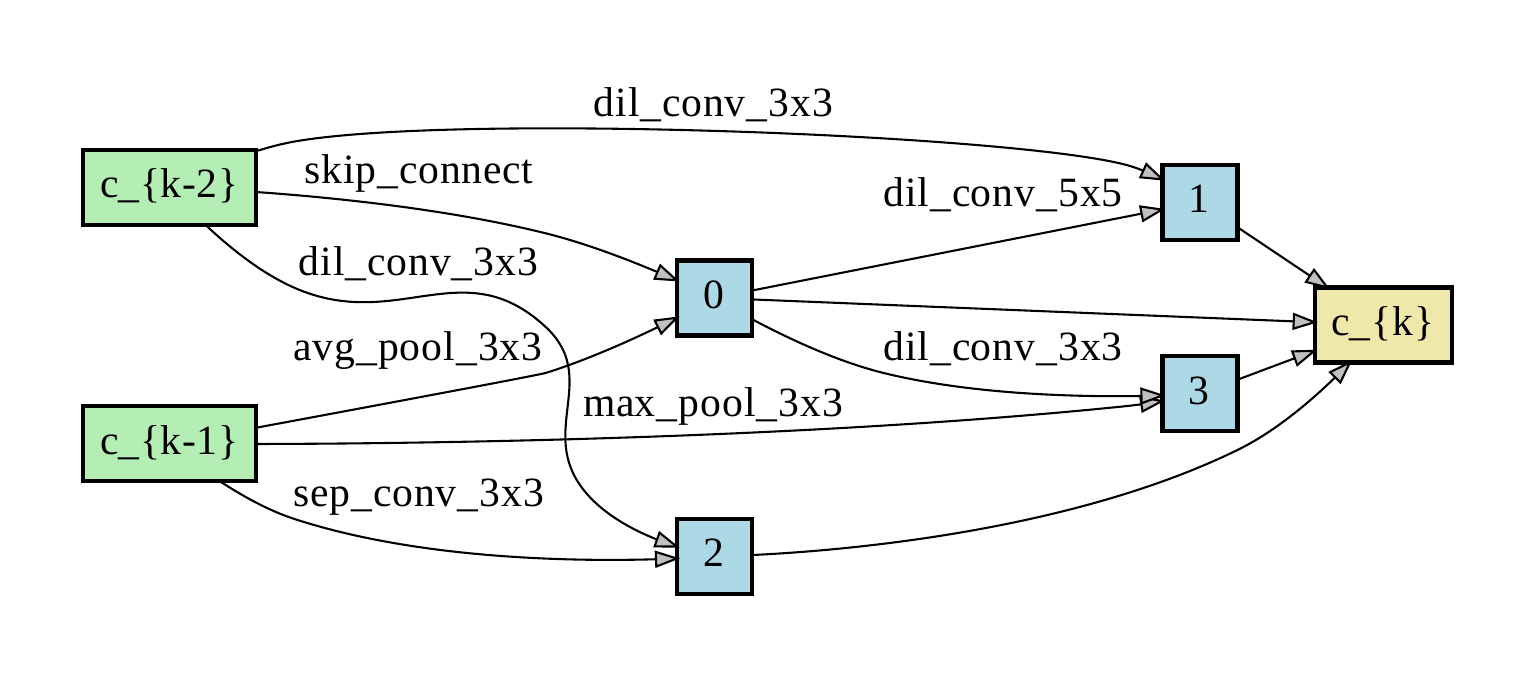}
    \includegraphics[width = 0.25\textwidth]{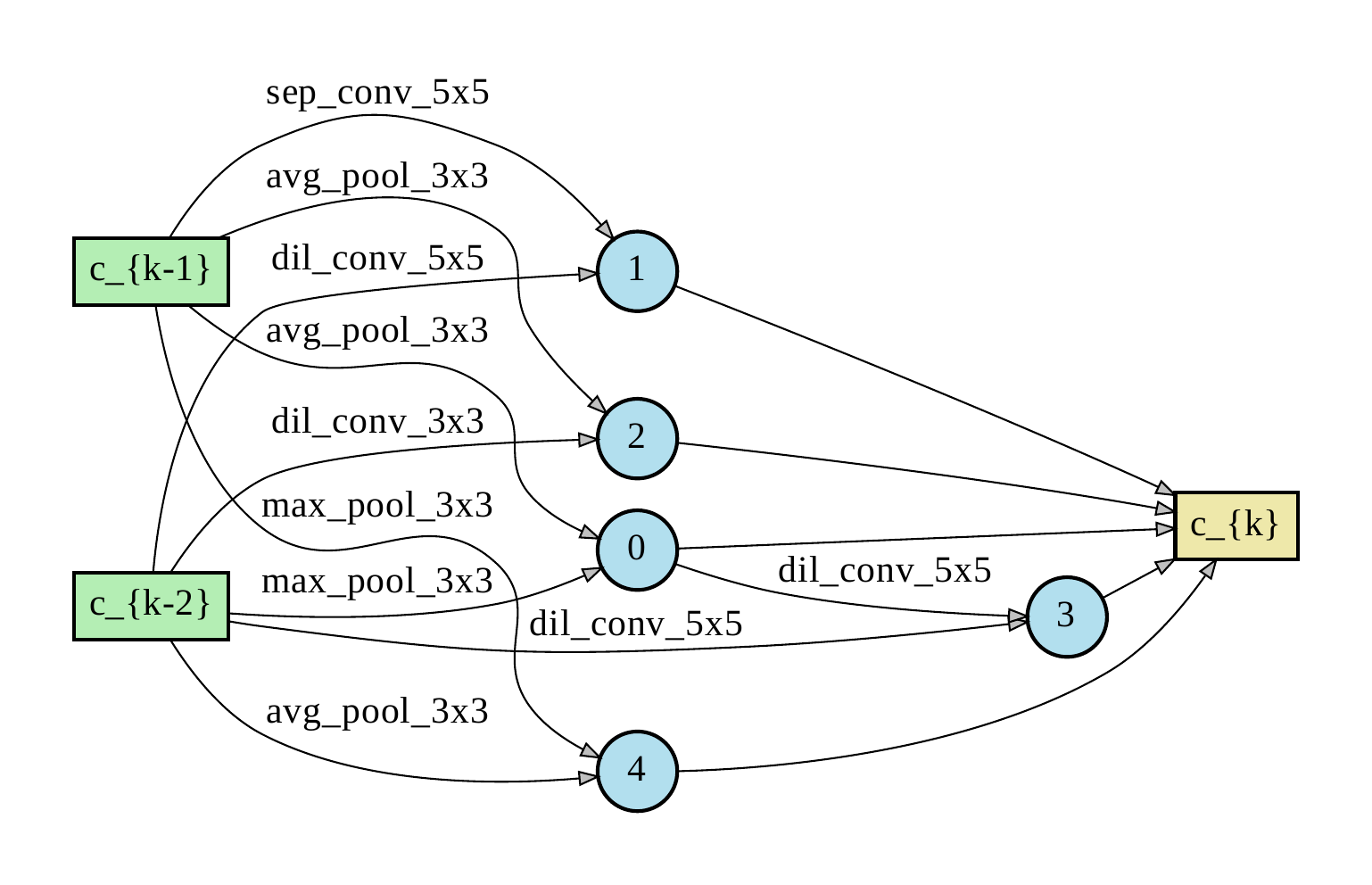}
    \includegraphics[width = 0.2\textwidth]{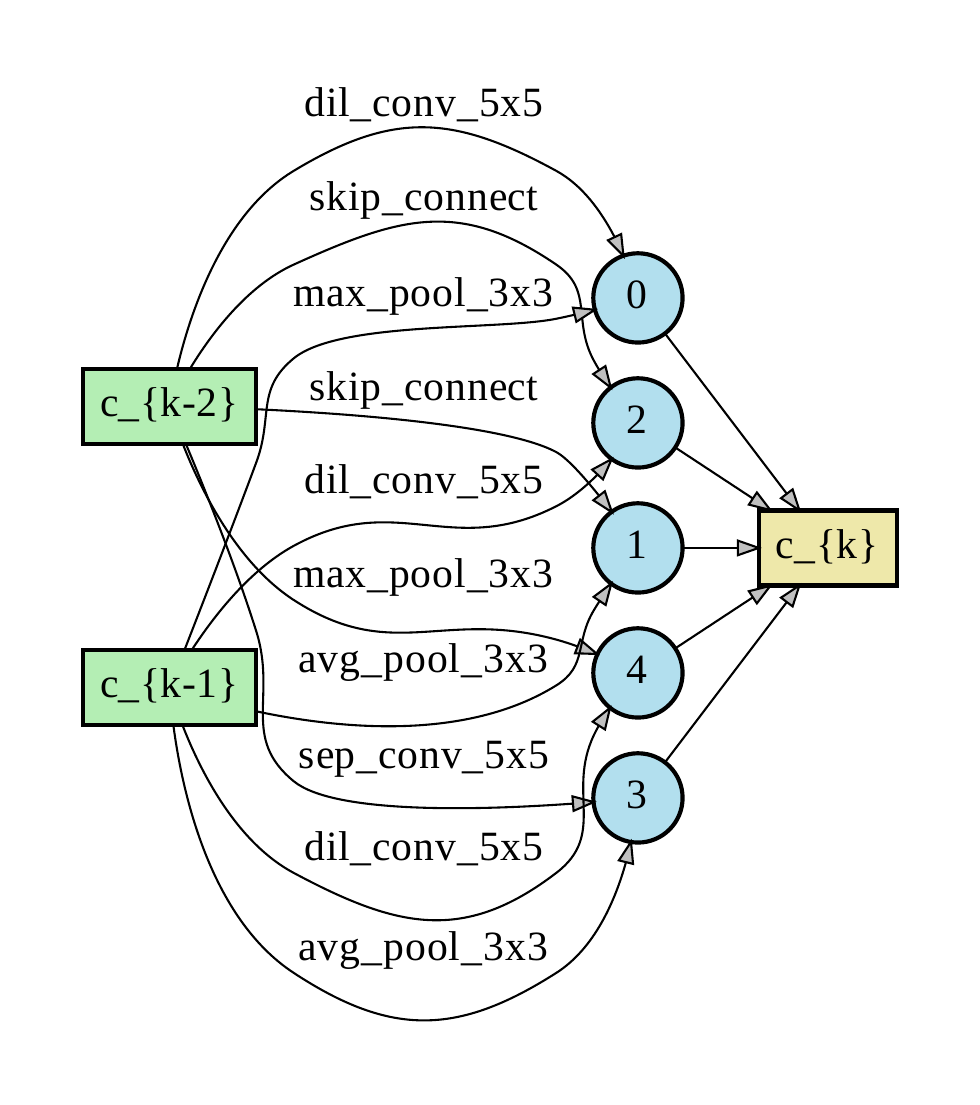}
    \caption{Cell structures found our proposed approach: Top panel shows discovered normal (left) and reduced (right) cell architectures from search space associated to $\text{DermoCSNAS}_{N=4}$, while bottom panel are associated to $\text{DermoCSNAS}_{N=5}$.}
    \label{c3}
\end{figure}

We report neural cell discovered by CSNAS in Figure~\ref{c1}, ~\ref{c2} and ~\ref{c3}.


\subsection{Neural Architecture Validation} \label{appendix:arceval}

We follow the setup used in \cite{zoph2018learning,liu2018darts,liu2018progressive,real2019regularized}, where the first and second nodes of cell $C_{k}$ are set to be compatible to the outputs of cell $C_{k-2}$ and $C_{k-1}$, respectively. All reduction cells are located in 1/2 and 2/3 of the depth, which has stride equals two for all operations linked to the input node.

We construct a large network for the CIFAR-10 dataset, including $20$ layers with $36$ initial channels. The train setting is employed from existing studies \cite{zoph2016neural,zoph2018learning,liu2018darts,liu2018progressive,real2019regularized}, offering more regularization on training, which includes: cutout \cite{devries2017improved} of length 16, linearly path drop out with probability $0.2$ and auxiliary classifier (located in 2/3 maximum depth of the network) with weight $0.4$. We train the network for $600$ epochs using batch size $128$. The chosen optimizer is momentum SGD with learning rate $\eta = 0.025$, momentum = $0.9$, weights decay $3\times 10^{-4}$ and gradient clip of $5$. The entire training process takes three days on one single GPU.

Regarding the ImageNet dataset, the input resolution is set to be $224 \times 224$, and the allowed number of multiply-add operations is less than 600 number of operations, which is restricted for mobile settings. We train a network having $14$ cells and $48$ initial channels for $250$ epochs with a batch size of $128$. The learning rate is set at $\eta = 0.1$ with decay rate $0.97$ and decay period of $2.5$. We use the same auxiliary module as the evaluation phase of CIFAR-10. We set the SGD optimizer at a momentum of $0.9$, and weights decay $3 \times 10^{-5}$.
\subsection{Hyper-parameters setting for comparison between UnNAS and CSNAS}\label{appendix:unnas-csnas}

This section reports the evaluation setting for comparison between UnNAS and CSNAS. The model includes $14$ layers. We train the network for $250$ epochs with initial learning rate of $0.5$ under SGD optimizer. The number of warm-up epoch is $5$ and cosine learning rate schedule is used. Moreover, we use batch size of $512$ distributed over $4$ GPUs. The weight for auxiliary loss is $0.4$ and the drop-path-probability is 0.3.

\subsection{Visualization of CAM}
\begin{figure}[ht]
    \centering
    \includegraphics[clip, trim=0cm 5cm 0cm 0cm,width = 0.24\textwidth]{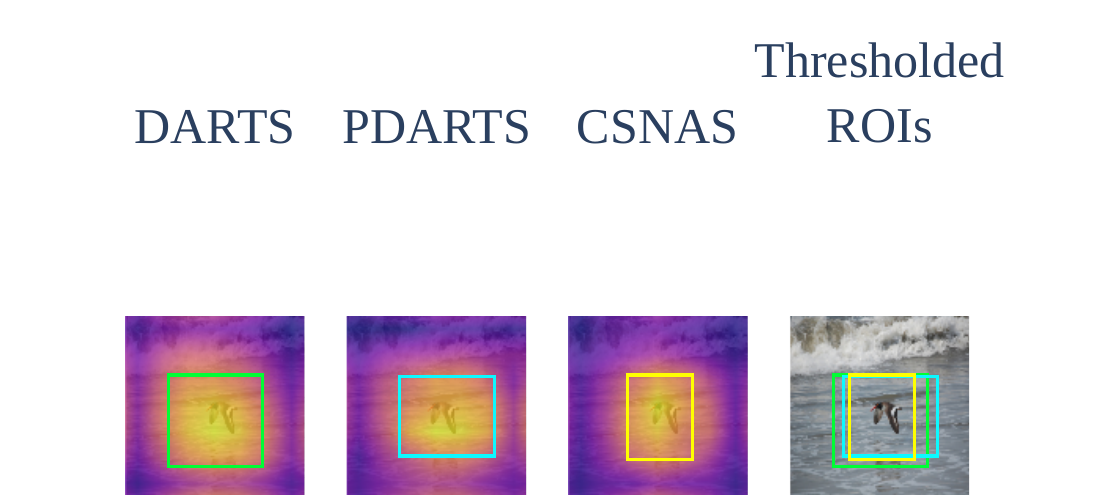}
    \includegraphics[clip, trim=0cm 5cm 0cm 5cm, width = 0.24\textwidth]{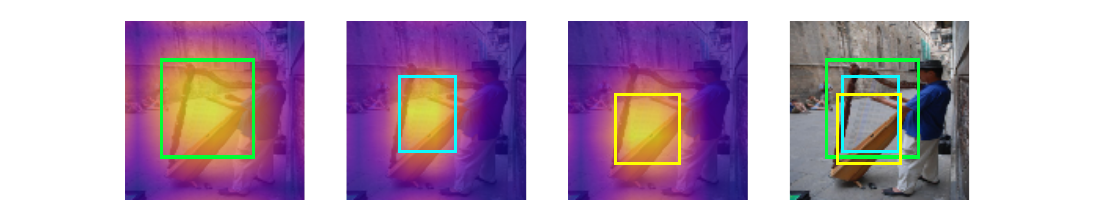}
    \includegraphics[clip, trim=0cm 5cm 0cm 5cm, width = 0.24\textwidth]{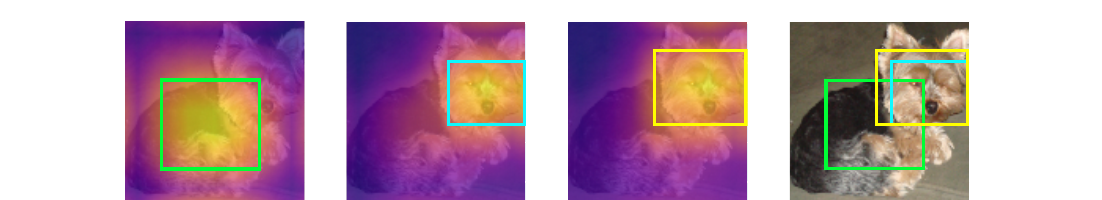}
    \includegraphics[clip, trim=0cm 5cm 0cm 5cm, width = 0.24\textwidth]{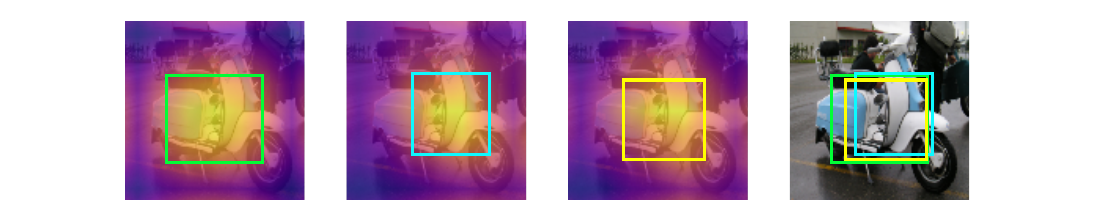}
    \includegraphics[clip, trim=0cm 5cm 0cm 5cm, width = 0.24\textwidth]{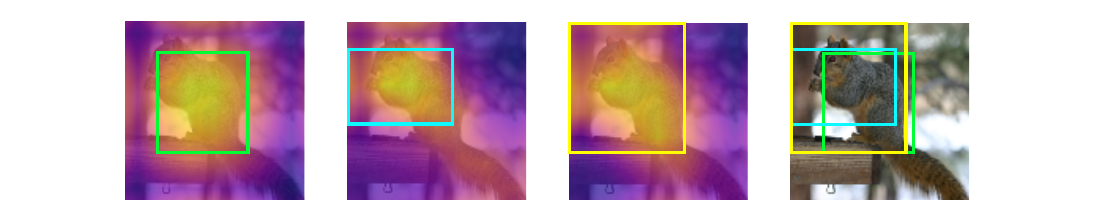}
    \includegraphics[clip, trim=0cm 5cm 0cm 5cm, width = 0.24\textwidth]{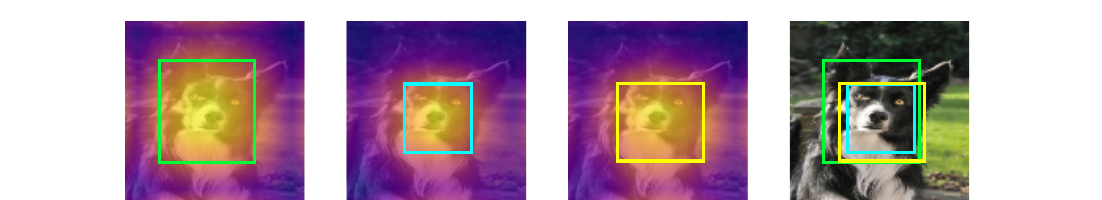}
    \includegraphics[clip, trim=0cm 5cm 0cm 5cm, width = 0.24\textwidth]{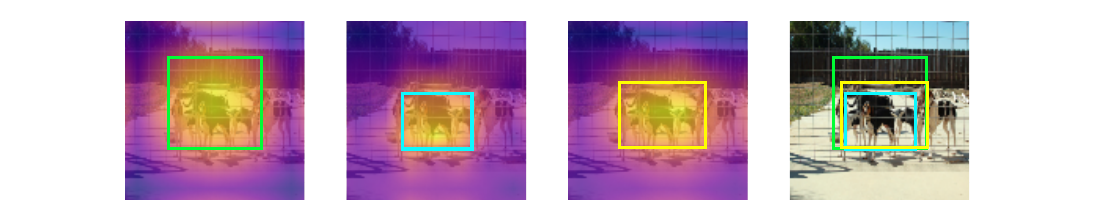}
    \includegraphics[clip, trim=0cm 5cm 0cm 5cm, width = 0.24\textwidth]{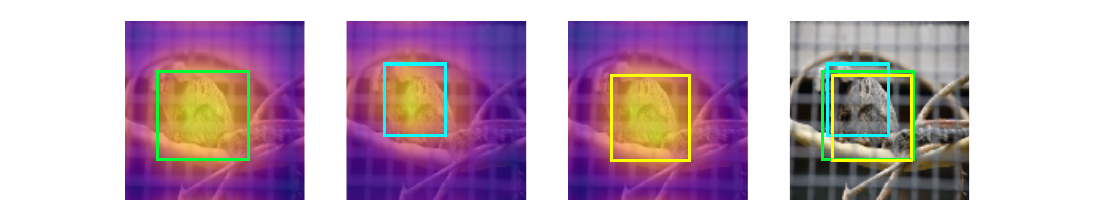}
    \includegraphics[clip, trim=0cm 5cm 0cm 5cm, width = 0.24\textwidth]{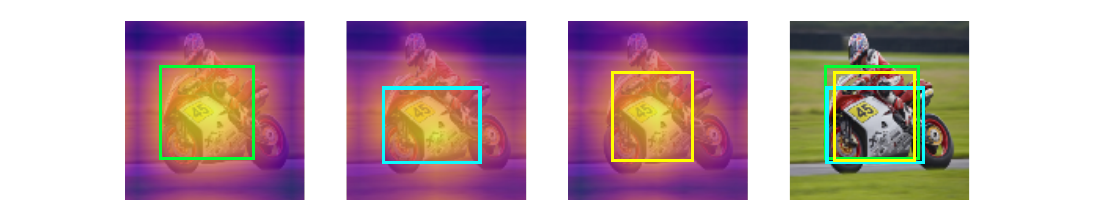}
    \includegraphics[clip, trim=0cm 5cm 0cm 5cm, width = 0.24\textwidth]{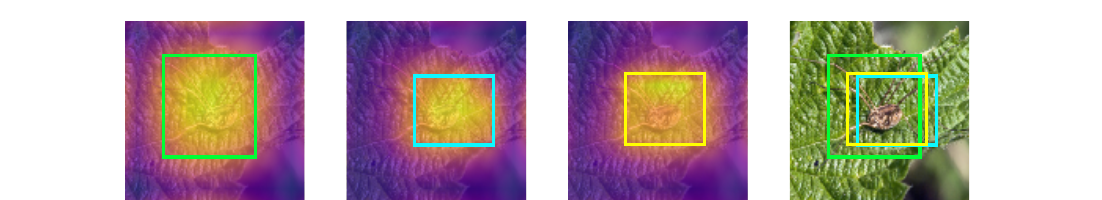}

    \caption{\textbf{Visualization of CAM.} In the first three columns, we represent the Classification Activation Maps (CAM) of ten random samples from DARTS, PDARTS and CSNAS. The last column depicts the thresholded ROIs from CAMs with threshold of $\tau=0.7$.}
    \label{cam_imgnet}
\end{figure}



\begin{IEEEbiography}[{\includegraphics[width=1in,height=1.25in,clip,keepaspectratio]{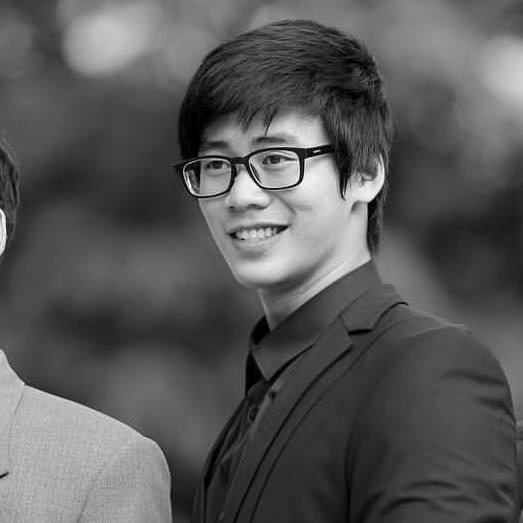}}]{Nam Nguyen} Nam Nguyen received B.S (2018) in Mathematics from Hanoi University of Education, M.A (2019) in Statistics from University of South Florida. Recently, he is a graduate student in Electrical Engineering at the University of South Florida. His research interest is neural architecture design and representation learning.
\end{IEEEbiography}

\begin{IEEEbiography}[{\includegraphics[width=1in,height=1.25in,clip,keepaspectratio]{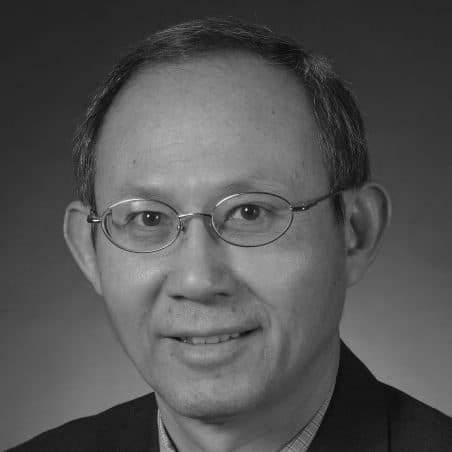}}]{J. Morris Chang} J. Morris Chang received the BSEE degree from the Tatung Institute of Technology, Taiwan, and the MS and Ph.D. degrees in computer engineering from North Carolina State University. He is currently a professor with the Department of Electrical Engineering, University of South Florida. His industrial experience includes positions at Texas Instruments, Taiwan, Microelectronics Center of North Carolina, and AT\&T Bell Laboratories, Pennsylvania. He was on the Department of Electrical Engineering, Rochester Institute of Technology, Rochester, the Department of Computer Science, Illinois Institute of Technology, Chicago, and the Department of Electrical and Computer Engineering, Iowa State University, IA. His research interests include cybersecurity, wireless networks, energy-aware computing, and object-oriented systems. Currently, he is a handling editor of the Journal of Microprocessors and Microsystems and the IEEE IT Professional associate editor-in-chief. He is a senior member of the IEEE. (Based on a document published on 18 July 2019).
\end{IEEEbiography}

\end{document}